\documentclass[11pt,a4paper]{article}
\usepackage[utf8]{inputenc}
\usepackage{amsmath,amssymb}
\usepackage{array}
\usepackage{graphicx}
\usepackage[hidelinks]{hyperref}
\usepackage{booktabs}
\usepackage[margin=1in]{geometry}
\usepackage{natbib}
\usepackage{subcaption}
\usepackage{xcolor}
\usepackage{microtype}
\usepackage{url}

\newcommand{\NumLanguages}{135}
\newcommand{\NumConcepts}{101}
\newcommand{\SwadeshMean}{0.58}
\newcommand{\SwadeshStd}{0.17}
\newcommand{\SwadeshMax}{0.91}
\newcommand{\SwadeshMin}{0.12}
\newcommand{\SwadeshTopConcept}{night}
\newcommand{\SwadeshBottomConcept}{louse}
\newcommand{\MantelRho}{0.13}
\newcommand{\MantelP}{0.020}
\newcommand{\MantelPermutations}{999}
\newcommand{\MantelNumLangs}{88}
\newcommand{\MantelWithinSubfamilyRho}{0.65}
\newcommand{\MantelWithinSubfamilyP}{2.40 \times 10^{-40}}
\newcommand{\MantelWithinSubfamilyPairs}{321}
\newcommand{\MantelCrossBranchIERho}{0.10}
\newcommand{\MantelCrossBranchIEPairs}{374}
\newcommand{\MantelBetweenFamilyRho}{0.04}
\newcommand{\MantelBetweenFamilyPairs}{3133}
\newcommand{\MantelTotalPairs}{3828}
\newcommand{\MantelBetweenFamilyPct}{81.8}
\newcommand{\MantelWithinSubfamilyBlockP}{1.00 \times 10^{-4}}
\newcommand{\MantelBareWithinSubfamilyRho}{0.57}
\newcommand{\MantelBareWithinSubfamilyBlockP}{2.00 \times 10^{-4}}
\newcommand{\MantelCrossScriptSubfamilyPairs}{53}
\newcommand{\MantelCrossScriptSubfamilyRho}{0.57}
\newcommand{\MantelCrossScriptSubfamilyP}{6.95 \times 10^{-6}}
\newcommand{\MantelBareCrossScriptSubfamilyRho}{0.60}
\newcommand{\MantelBareCrossScriptSubfamilyP}{2.41 \times 10^{-6}}
\newcommand{\MantelWithinSubfamilyPartialRho}{0.67}
\newcommand{\MantelLDNDSubfamilyRho}{0.67}
\newcommand{\MantelLDNDRho}{0.15}
\newcommand{\RHMHighResCount}{40}
\newcommand{\RHMLowResCount}{95}
\newcommand{\RHMHighResDist}{0.24}
\newcommand{\RHMLowResDist}{0.30}
\newcommand{\RHMResourceU}{690}
\newcommand{\RHMResourceP}{2.80 \times 10^{-9}}
\newcommand{\MantelBareRho}{0.24}
\newcommand{\MantelBareP}{0.001}
\newcommand{\PhyloNumLanguages}{135}
\newcommand{\SwadeshCompMean}{0.80}
\newcommand{\NonSwadeshCompMean}{0.89}

\newcommand{\SwadeshCompNumNonSwadesh}{60}
\newcommand{\OrthoSimLoanword}{0.26}
\newcommand{\ControlledSwadeshCompMean}{0.80}
\newcommand{\ControlledNonSwadeshCompMean}{0.78}
\newcommand{\ControlledSwadeshCompU}{3420}
\newcommand{\ControlledSwadeshCompP}{0.087}
\newcommand{\ControlledSwadeshCompCohenD}{0.23}
\newcommand{\ControlledSwadeshCompNumSwadesh}{101}
\newcommand{\ControlledSwadeshCompNumNonSwadesh}{60}
\newcommand{\PosNounSwadeshCount}{52}
\newcommand{\PosNounSwadeshMean}{0.83}
\newcommand{\PosNounNonSwadeshMean}{0.78}
\newcommand{\PosNounCompU}{2016}
\newcommand{\PosNounCompP}{0.004}
\newcommand{\PosNounCompCohenD}{0.45}
\newcommand{\PosNounCorrSwadeshMean}{0.75}
\newcommand{\PosNounCorrNonSwadeshMean}{0.65}
\newcommand{\PosNounCorrCompU}{2099}
\newcommand{\PosNounCorrCompP}{8.40 \times 10^{-4}}
\newcommand{\PosNounCorrCompCohenD}{0.54}
\newcommand{\OrthoSimSwadesh}{0.13}
\newcommand{\OrthoSimSwadeshNouns}{0.12}
\newcommand{\OrthoSimControlled}{0.13}
\newcommand{\StrictNounCount}{52}
\newcommand{\StrictNounRawNonSwadeshMean}{0.76}
\newcommand{\StrictNounRawU}{1846}
\newcommand{\StrictNounRawP}{7.16 \times 10^{-4}}
\newcommand{\StrictNounRawCohenD}{0.59}
\newcommand{\StrictNounCorrNonSwadeshMean}{0.63}
\newcommand{\StrictNounCorrU}{1933}
\newcommand{\StrictNounCorrP}{8.03 \times 10^{-5}}
\newcommand{\StrictNounCorrCohenD}{0.69}
\newcommand{\SwadeshNounMeanFert}{1.73}
\newcommand{\NonSwadeshNounMeanFert}{2.21}
\newcommand{\SwadeshNounMeanLen}{4.19}
\newcommand{\NonSwadeshNounMeanLen}{4.85}
\newcommand{\NounLenDiffP}{9.02 \times 10^{-9}}
\newcommand{\NounFertAncovaT}{-5.26}
\newcommand{\NounFertAncovaP}{7.43 \times 10^{-7}}
\newcommand{\NounSwadeshAncovaP}{0.510}
\newcommand{\ColexSpearmanRho}{0.13}
\newcommand{\ColexSpearmanP}{8.85 \times 10^{-21}}
\newcommand{\ColexNumPairs}{5050}
\newcommand{\ColexU}{492477}
\newcommand{\ColexP}{1.55 \times 10^{-33}}
\newcommand{\ColexColMean}{0.37}
\newcommand{\ColexNonColMean}{0.29}
\newcommand{\ColexCohenD}{1.43}
\newcommand{\ColexColCount}{201}
\newcommand{\ColexNonColCount}{3258}

\newcommand{\ColexStringDistinctColMean}{0.37}
\newcommand{\ColexStringDistinctNonColMean}{0.29}
\newcommand{\ColexStringDistinctU}{492111}
\newcommand{\ColexStringDistinctP}{2.14 \times 10^{-33}}
\newcommand{\ColexStringDistinctCohenD}{1.41}
\newcommand{\ColexStringDistinctRho}{0.13}
\newcommand{\ColexStringDistinctRhoP}{1.19 \times 10^{-20}}
\newcommand{\ColexWithinCatPairs}{783}
\newcommand{\ColexWithinCatColMean}{0.42}
\newcommand{\ColexWithinCatNonColMean}{0.33}
\newcommand{\ColexWithinCatU}{27839}
\newcommand{\ColexWithinCatP}{4.51 \times 10^{-18}}
\newcommand{\ColexWithinCatCohenD}{1.26}
\newcommand{\ColexWithinCatRho}{0.32}
\newcommand{\ColexWithinCatRhoP}{2.73 \times 10^{-20}}
\newcommand{\ColexSamePOSPairs}{1765}
\newcommand{\ColexSamePOSColMean}{0.39}
\newcommand{\ColexSamePOSNonColMean}{0.31}
\newcommand{\ColexSamePOSU}{101245}
\newcommand{\ColexSamePOSP}{2.10 \times 10^{-22}}
\newcommand{\ColexSamePOSCohenD}{1.20}
\newcommand{\ColexSamePOSRho}{0.20}

\newcommand{\ColexBareStringDistinctU}{528369}
\newcommand{\ColexBareStringDistinctP}{9.98 \times 10^{-49}}
\newcommand{\ColexBareStringDistinctCohenD}{1.79}
\newcommand{\ColexBareStringDistinctRho}{0.22}
\newcommand{\ColexBareStringDistinctRhoP}{6.95 \times 10^{-55}}
\newcommand{\ColexCentroidCohenD}{1.80}
\newcommand{\ColexCentroidRho}{0.20}
\newcommand{\ColexCentroidRhoP}{5.91 \times 10^{-47}}
\newcommand{\ColexBareCentroidCohenD}{1.71}
\newcommand{\ColexBareCentroidRho}{0.21}
\newcommand{\ColexBareCentroidRhoP}{2.68 \times 10^{-53}}
\newcommand{\ColexWithinCatColexCount}{399}
\newcommand{\ColexCrossCatColexCount}{1{,}393}
\newcommand{\ColexCognateExclU}{491{,}254}
\newcommand{\ColexCognateExclP}{3.55 \times 10^{-33}}
\newcommand{\ColexCognateExclCohenD}{1.40}
\newcommand{\ColexOneFamCount}{1{,}241}
\newcommand{\ColexOneFamMean}{0.30}
\newcommand{\ColexFiveFamCount}{63}
\newcommand{\ColexFiveFamMean}{0.44}
\newcommand{\ColexTenFamCount}{31}
\newcommand{\ColexTenFamMean}{0.48}
\newcommand{\ConceptualStoreRaw}{2.25}
\newcommand{\ConceptualStoreCentered}{2.69}
\newcommand{\ConceptualStoreImprovement}{1.19}
\newcommand{\ConceptualStoreBareRaw}{1.15}
\newcommand{\ConceptualStoreBareCentered}{2.46}
\newcommand{\ConceptualStoreBareImprovement}{2.14}
\newcommand{\CSMSSConceptCtx}{64.4}
\newcommand{\CSMSSLangCtx}{3.8}
\newcommand{\CSMSSResidCtx}{31.8}
\newcommand{\CSMSSConceptBare}{44.7}
\newcommand{\CSMSSLangBare}{25.7}
\newcommand{\CSMSSResidBare}{29.6}
\newcommand{\CSMPermNull}{1.01}
\newcommand{\CSMBareGlobalCentered}{1.76}
\newcommand{\CSMCrossScriptCtxRaw}{2.47}
\newcommand{\CSMCrossScriptCtxCentered}{2.94}
\newcommand{\CSMCrossScriptCtxImprovement}{1.19}
\newcommand{\CSMCrossScriptBareRaw}{1.15}
\newcommand{\CSMCrossScriptBareCentered}{2.60}
\newcommand{\CSMCrossScriptBareImprovement}{2.27}
\newcommand{\PRLangBare}{13.3}
\newcommand{\PRLangCtx}{18.8}
\newcommand{\PRConcept}{38.7}
\newcommand{\PRLangBareTopTenPct}{66.6}
\newcommand{\PRConceptTopTenPct}{40.1}
\newcommand{\OffsetNumPairs}{22}
\newcommand{\OffsetMeanConsistency}{0.84}
\newcommand{\OffsetMaxConsistency}{0.94}
\newcommand{\OffsetMinConsistency}{0.70}
\newcommand{\OffsetBestPair}{fire--water}
\newcommand{\OffsetNormCorrR}{0.84}
\newcommand{\OffsetNormCorrP}{1.03 \times 10^{-6}}
\newcommand{\OffsetSemanticCount}{15}
\newcommand{\OffsetControlCount}{7}
\newcommand{\OffsetSemanticMean}{0.82}
\newcommand{\OffsetControlMean}{0.86}

\newcommand{\OffsetBestSemanticVal}{0.94}
\newcommand{\OffsetResidSemanticMean}{0.012}
\newcommand{\OffsetResidControlMean}{-0.025}
\newcommand{\OffsetResidU}{90}
\newcommand{\OffsetResidP}{0.003}
\newcommand{\OffsetResidCohenD}{1.06}
\newcommand{\OffsetBestResidPair}{man--woman}
\newcommand{\OffsetCouplingSemanticMean}{0.17}
\newcommand{\OffsetCouplingControlMean}{-0.00}
\newcommand{\OffsetCouplingU}{105}
\newcommand{\OffsetCouplingP}{5.86 \times 10^{-6}}
\newcommand{\OffsetLOOU}{89}
\newcommand{\OffsetLOOP}{0.004}
\newcommand{\OffsetConcaveU}{83}
\newcommand{\OffsetConcaveP}{0.016}
\newcommand{\OffsetBareConcaveU}{96}
\newcommand{\OffsetBareConcaveP}{6.00 \times 10^{-4}}
\newcommand{\OffsetBareCouplingSemMean}{0.23}
\newcommand{\OffsetBareCouplingCtrlMean}{-0.01}
\newcommand{\OffsetBareCouplingU}{105}
\newcommand{\OffsetBareCouplingP}{5.86 \times 10^{-6}}
\newcommand{\OffsetMatchedCount}{15}
\newcommand{\OffsetMatchedSemMean}{0.82}
\newcommand{\OffsetMatchedCtrlMean}{0.78}
\newcommand{\OffsetMatchedRawU}{154}
\newcommand{\OffsetMatchedRawP}{0.045}
\newcommand{\OffsetMatchedResidU}{211}
\newcommand{\OffsetMatchedResidP}{3.55 \times 10^{-5}}
\newcommand{\OffsetMatchedCouplingU}{225}
\newcommand{\OffsetMatchedCouplingP}{1.70 \times 10^{-6}}
\newcommand{\OffsetMatchedSemNormMean}{10.18}
\newcommand{\OffsetMatchedCtrlNormMean}{10.74}
\newcommand{\OffsetMatchedNormP}{0.30}
\newcommand{\RHMHighResOffset}{0.89}
\newcommand{\RHMLowResOffset}{0.80}
\newcommand{\RHMOffsetU}{3{,}101}
\newcommand{\RHMOffsetP}{3.63 \times 10^{-9}}
\newcommand{\ColorNumColors}{11}
\newcommand{\ColorNumLanguages}{135}
\newcommand{\ColorPConeVar}{22.8}
\newcommand{\ColorPCtwoVar}{16.3}
\newcommand{\ColorPCthreeVar}{13.5}
\newcommand{\ColorPConeTwoVar}{39.1}
\newcommand{\ColorTotalVar}{52.6}
\newcommand{\ColorBKRho}{0.93}
\newcommand{\ColorBKP}{8.48 \times 10^{-5}}
\newcommand{\ColorBKR}{0.89}
\newcommand{\ColorBKRp}{2.83 \times 10^{-4}}
\newcommand{\ColorBKRhoAll}{0.86}
\newcommand{\ColorBKPAll}{6.08 \times 10^{-4}}
\newcommand{\ColorBKRSeven}{0.82}
\newcommand{\ColorBKRSevenP}{0.002}
\newcommand{\ColorBarePConeVar}{23.8}
\newcommand{\ColorBareBKRho}{0.91}
\newcommand{\ColorBareBKP}{2.18 \times 10^{-4}}
\newcommand{\ColorBareBKR}{0.88}
\newcommand{\ColorBareBKRp}{4.18 \times 10^{-4}}
\newcommand{\ColorPartialLengthR}{0.70}
\newcommand{\ColorPartialLengthP}{0.017}
\newcommand{\ColorBarePartialLengthR}{0.63}
\newcommand{\ColorBarePartialLengthP}{0.038}
\newcommand{\ColorLayerZeroRho}{0.81}
\newcommand{\ColorLayerZeroP}{0.0046}
\newcommand{\ColorLayerSevenRho}{0.96}
\newcommand{\ColorLayerSevenP}{7.69 \times 10^{-6}}
\newcommand{\RHMHighResColorDist}{1.42}
\newcommand{\RHMLowResColorDist}{2.05}
\newcommand{\RHMColorP}{1.65 \times 10^{-8}}
\newcommand{\IsotropySpearmanRho}{0.990}
\newcommand{\IsotropySpearmanP}{4.93 \times 10^{-87}}
\newcommand{\DecompRsqOrtho}{0.012}
\newcommand{\DecompSlopeOrtho}{1.308}
\newcommand{\DecompPOrtho}{0.272}
\newcommand{\DecompRsqPhon}{0.000}
\newcommand{\DecompSlopePhon}{0.098}
\newcommand{\DecompPPhon}{0.929}
\newcommand{\DecompNumLatinLangs}{92}
\newcommand{\CatBodyMean}{0.55}
\newcommand{\CatBodyStd}{0.17}
\newcommand{\CatNatureMean}{0.72}
\newcommand{\CatNatureStd}{0.15}
\newcommand{\CatAnimalsMean}{0.60}
\newcommand{\CatAnimalsStd}{0.24}
\newcommand{\CatPeopleMean}{0.77}
\newcommand{\CatPeopleStd}{0.05}
\newcommand{\CatActionsMean}{0.54}
\newcommand{\CatActionsStd}{0.13}
\newcommand{\CatPropertiesMean}{0.57}
\newcommand{\CatPropertiesStd}{0.11}
\newcommand{\CatPronounsMean}{0.45}
\newcommand{\CatPronounsStd}{0.08}
\newcommand{\CatOtherMean}{0.54}
\newcommand{\CatOtherStd}{0.20}
\newcommand{\CatAnovaF}{4.53}
\newcommand{\CatAnovaP}{2.21 \times 10^{-4}}

\newcommand{\CatAnovaEtaPct}{25.4}
\newcommand{\CatKWH}{28.00}
\newcommand{\CatKWP}{2.20 \times 10^{-4}}
\newcommand{\CatBareAnovaF}{3.00}
\newcommand{\CatBareAnovaP}{0.007}
\newcommand{\CatBareAnovaEtaPct}{18.4}
\newcommand{\CatPartialFertF}{7.74}
\newcommand{\CatPartialFertP}{2.51 \times 10^{-7}}
\newcommand{\DecompRsqFert}{0.037}
\newcommand{\DecompPFert}{0.055}
\newcommand{\DecompRsqLen}{0.000}
\newcommand{\DecompPLen}{0.837}
\newcommand{\CarrierCtxMean}{0.58}
\newcommand{\CarrierDectxMean}{0.45}
\newcommand{\CarrierBaselineRho}{0.866}
\newcommand{\CarrierBaselineP}{1.48 \times 10^{-31}}
\newcommand{\CarrierBaselineMeanDiff}{0.129}
\newcommand{\CarrierBaselineTstat}{15.74}
\newcommand{\CarrierBaselineTp}{7.97 \times 10^{-29}}
\newcommand{\LayerwiseNumLayers}{12}
\newcommand{\LayerwiseNumLangs}{39}
\newcommand{\LayerwiseEmergenceLayer}{1}
\newcommand{\LayerwisePeakLayer}{10}
\newcommand{\LayerwisePeakConv}{0.86}
\newcommand{\LayerwisePhaseTrans}{6}
\newcommand{\LayerwiseFinalCSM}{4.01}
\newcommand{\LayerwiseFinalConv}{0.80}
\newcommand{\LayerwiseInputConv}{0.35}
\newcommand{\LayerTenCenteredCSM}{3.89}
\newcommand{\LayerSixRawBetween}{0.466}
\newcommand{\LayerTenRawBetween}{0.462}
\newcommand{\LayerElevenRawBetween}{0.680}
\newcommand{\LayerBetweenGainPct}{47.1}
\newcommand{\LayerWithinContractPct}{43.3}
\newcommand{\LayerWithinExpandPct}{45.5}
\newcommand{\LayerPronounDip}{-0.15}
\newcommand{\LayerActionDip}{-0.09}

\newcommand{\IsotropyMinRho}{0.98}
\newcommand{\IsotropyMaxRho}{1.00}
\newcommand{\IsotropyKRange}{0.98\text{--}1.00}
\newcommand{\IsotropyKZeroMean}{0.72}
\newcommand{\IsotropyKOneMean}{0.59}
\newcommand{\IsotropyKThreeMean}{0.58}
\newcommand{\IsotropyKTenMean}{0.56}

\begin{document}
\title{Universal Conceptual Structure in Neural Translation:\\
Probing NLLB-200's Multilingual Geometry}

\author{
  Kyle Mathewson \\
  University of Alberta \\
  \texttt{kyle.mathewson@ualberta.ca}
}

\date{\today}

\maketitle

\begin{abstract}
Do multilingual translation models organize meaning around a shared conceptual space, or do they primarily encode language identity and surface form?
We probe the encoder geometry of Meta's 600M-parameter distilled NLLB-200 translation model (\texttt{facebook/\allowbreak nllb-200-\allowbreak distilled-600M}) across \NumLanguages{} typologically diverse languages and \NumConcepts{} Swadesh core concepts, evaluating both target words embedded in a shared English carrier frame and decontextualized native bare-word inputs against targeted null and confound controls.
First, two-way variance partitioning shows that NLLB-200's encoder implements an additive language-shift plus conceptual-manifold geometry ($\mathbf{e}_{c,\ell} = \bar{\mathbf{e}} + \hat{\boldsymbol{\mu}}_c + \hat{\boldsymbol{\ell}} + \hat{\boldsymbol{\epsilon}}_{c,\ell}$): per-language shifts $\hat{\boldsymbol{\ell}}$ account for $\CSMSSLangBare\%$ of total variance on native bare words versus $\CSMSSLangCtx\%$ in the shared carrier frame, and subtracting $\hat{\boldsymbol{\ell}}$ increases the between-concept to within-concept distance ratio (CSM, vs.\ a $\CSMPermNull$ permutation null) by $\ConceptualStoreBareImprovement\times$ on bare words ($\ConceptualStoreBareRaw \to \ConceptualStoreBareCentered$; $\CSMCrossScriptBareImprovement\times$ across strictly disjoint scripts) and by $\ConceptualStoreImprovement\times$ in the carrier frame ($\ConceptualStoreRaw \to \ConceptualStoreCentered$), with conceptual separation emerging across intermediate encoder layers (inflection at Layer~\LayerwisePhaseTrans{}, convergence peaking at Layer~\LayerwisePeakLayer{}).
Second, semantic category explains $\eta^2 = \CatAnovaEtaPct\%$ of cross-lingual convergence variance across the Swadesh list ($F(7,93) = \CatAnovaF$, $p = \CatAnovaP$; partial $F = \CatPartialFertF$, $p = \CatPartialFertP$ controlling for subword fertility and word length)—over twenty times the variance explained by orthographic or phonological similarity ($R^2 \le \DecompRsqOrtho$)—while the higher convergence of core Swadesh nouns over non-loanword concrete nouns ($d = \PosNounCorrCompCohenD$--$\StrictNounCorrCohenD$, $p \le \PosNounCorrCompP$) is mechanistically mediated by Zipfian subword compactness ($p = \NounFertAncovaP$).
Third, external typological benchmarks confirm that the conceptual manifold reflects both genealogical history and universal human semantic associations: language-pair distances track ASJP phonetic distances within language subfamilies under block-permutation and cross-script controls ($\rho_s = \MantelWithinSubfamilyRho$, block $p = \MantelWithinSubfamilyBlockP$; cross-script $\rho_s = \MantelCrossScriptSubfamilyRho$--$\MantelBareCrossScriptSubfamilyRho$; global bare-word Mantel $\rho_s = \MantelBareRho$, $p = \MantelBareP$), and concept pairs frequently colexified in CLICS\textsuperscript{3} lie significantly closer in embedding space across string-distinct, within-category ($\rho_s = \ColexWithinCatRho$), and bare-word ($d = \ColexBareStringDistinctCohenD$) controls ($U = \ColexU$, $p = \ColexP$, $d = \ColexCohenD$).
Finally, the primary axis of basic color terms recovers the Berlin and Kay evolutionary hierarchy in both contextualized ($\rho_s = \ColorBKRho$, $p = \ColorBKP$) and bare-word ($\rho_s = \ColorBareBKRho$, $p = \ColorBareBKP$) regimes independently of word length, and semantic offset vectors exhibit significantly stronger distance-controlled residual alignment ($p = \OffsetResidP$, $d = \OffsetResidCohenD$) and doubly-centered perturbation coupling ($+\OffsetCouplingSemanticMean$ vs.\ $\OffsetCouplingControlMean$, $p = \OffsetCouplingP$) than negative controls.
We release InterpretCognates, an open-source interactive platform and reproducible evaluation suite for comparative representation geometry.
\end{abstract}

\section{Introduction}

Do neural machine translation models learn language-universal conceptual structure, or do they primarily organize representations around language identity and surface-form similarity?
This question sits at the intersection of mechanistic interpretability in natural language processing and a classic debate in cognitive science over whether multilingual speakers access a language-independent conceptual store or rely on language-specific lexical networks \citep{kroll1994,dijkstra2002,correia2014,deniz2025}.
Large-scale multilingual translation models offer a controlled computational testbed for this debate: because a single encoder must map hundreds of typologically diverse languages into a common representation space to support arbitrary many-to-many translation, its geometry reveals how statistical learning reconciles language-specific form with cross-linguistically invariant meaning.

Meta's NLLB-200 family \citep{nllbteam2022} was trained to translate directly among 200 languages, including dozens of low-resource languages across diverse scripts and language families.
In this study, we probe the 600-million-parameter distilled dense checkpoint (\texttt{facebook/nllb-200-distilled-600M}, 12 encoder layers, $d = 1024$), which compresses the translation capabilities of the 54.5B Mixture-of-Experts and 3.3B dense models into a compact shared encoder.
By evaluating \NumConcepts{} core concepts from the Swadesh list \citep{swadesh1952} across \NumLanguages{} languages after strict script and outlier quality filtering, and by contrasting words embedded in a shared English carrier frame against decontextualized native bare-word inputs, we test whether NLLB-200's encoder space can be decomposed into a grand mean $\bar{\mathbf{e}}$, an additive language shift $\boldsymbol{\ell}$, and a shared conceptual manifold $\boldsymbol{\mu}_c$ subject to zero-sum identifiability constraints ($\sum_c \boldsymbol{\mu}_c = \mathbf{0}$, $\sum_\ell \boldsymbol{\ell} = \mathbf{0}$):
\begin{equation}
\mathbf{e}_{c,\ell} \;=\; \bar{\mathbf{e}} \;+\; \boldsymbol{\mu}_c \;+\; \boldsymbol{\ell} \;+\; \boldsymbol{\epsilon}_{c,\ell}.
\label{eq:intro-decomposition}
\end{equation}

We organize our investigation into six progressive analyses that move from global representation geometry to external typological and domain-specific benchmarks, pairing each primary finding with targeted null and confound controls.
We begin in Section~\ref{sec:results-csm} by testing the additive language-shift model and tracing its emergence across the 12 encoder blocks.
Two-way sum-of-squares partitioning of Equation~\ref{eq:intro-decomposition} shows that while the shared conceptual manifold $\hat{\boldsymbol{\mu}}_c$ captures $\CSMSSConceptCtx{}\%$ of total variance in the carrier frame and $\CSMSSConceptBare{}\%$ on isolated bare words, per-language shifts $\hat{\boldsymbol{\ell}}$ account for $\CSMSSLangBare{}\%$ of total variance on native bare words versus $\CSMSSLangCtx{}\%$ in the shared carrier frame (concentrated in a low-dimensional language subspace of participation ratio $\text{PR} = \PRLangBare{}$ vs.\ $\PRConcept{}$ for concepts).
Subtracting $\hat{\boldsymbol{\ell}}$ increases the between-concept to within-concept distance ratio (Conceptual Store Metric, CSM; permutation null $= \CSMPermNull{}$) by $\ConceptualStoreBareImprovement{}\times$ on native bare-word inputs (from $\ConceptualStoreBareRaw{}$ to $\ConceptualStoreBareCentered{}$, and by $\CSMCrossScriptBareImprovement{}\times$ to $\CSMCrossScriptBareCentered{}$ across strictly disjoint writing systems) and by $\ConceptualStoreImprovement{}\times$ in the shared carrier frame (from $\ConceptualStoreRaw{}$ to $\ConceptualStoreCentered{}$), demonstrating that both regimes converge onto the same underlying conceptual store once language offsets are removed.
Across the 12 encoder blocks, cross-lingual convergence rises from $\LayerwiseInputConv{}$ at Layer~0 to peak at $\LayerwisePeakConv{}$ at Layer~\LayerwisePeakLayer{} before slightly specializing at Layer~11 ($\LayerwiseFinalConv{}$), while mean-centered CSM undergoes its sharpest inflection at Layer~\LayerwisePhaseTrans{}.

In Sections~\ref{sec:results-hierarchy} and~\ref{sec:results-confounds}, we examine the semantic hierarchy of cross-lingual convergence across the Swadesh lexicon and test whether it can be explained by surface-form or tokenization confounds.
After All-But-The-Top (ABTT) isotropy correction \citep{mu2018}, cross-lingual convergence across the \NumConcepts{} Swadesh concepts spans a wide, structured gradient ($\mu = \SwadeshMean{} \pm \SwadeshStd{}$), led by concrete human nouns and natural phenomena ($\mu = \CatPeopleMean{}$ and $\CatNatureMean{}$) and lowest for language-specific grammatical pronouns and deictics ($\mu = \CatPronounsMean{}$).
Semantic category explains $\eta^2 = \CatAnovaEtaPct{}\%$ of the variance in cross-lingual convergence ($F(7, 93) = \CatAnovaF{}$, $p = \CatAnovaP{}$), a result that strengthens after controlling for SentencePiece subword fertility and word length (partial $F(7, 91) = \CatPartialFertF{}$, $p = \CatPartialFertP{}$), is invariant to the number of principal components removed ($\rho_s \ge \IsotropyMinRho{}$ across $k \in \{0, 1, 3, 5, 10\}$), and persists on isolated native bare words ($\rho_s = \CarrierBaselineRho{}$, $p = \CarrierBaselineP{}$).
By contrast, across \DecompNumLatinLangs{} Latin-script languages, neither orthographic Levenshtein similarity ($R^2 = \DecompRsqOrtho{}$, $p = \DecompPOrtho{}$) nor normalized phonological similarity ($R^2 = \DecompRsqPhon{}$, $p = \DecompPPhon{}$) explains the Swadesh convergence hierarchy.
While modern international loanwords (\emph{coffee, radio, internet}) achieve artificially high raw similarity ($\mu = \NonSwadeshCompMean{}$) due to doubled orthographic overlap ($\OrthoSimLoanword{}$ vs.\ $\OrthoSimSwadesh{}$), matching Swadesh core concrete nouns ($n = \PosNounSwadeshCount{}$) against non-Swadesh concrete nouns ($n = \ControlledSwadeshCompNumNonSwadesh{}$, or $n = \StrictNounCount{}$ strictly tangible nouns) with identical low surface overlap ($\OrthoSimSwadeshNouns{}$ vs.\ $\OrthoSimControlled{}$) reveals significantly higher convergence for core Swadesh nouns ($p = \PosNounCompP{}$, $d = \PosNounCompCohenD{}$ raw; $p = \PosNounCorrCompP{}$, $d = \PosNounCorrCompCohenD{}$ ABTT-corrected; $d = \StrictNounCorrCohenD{}$ strictly concrete), which ANCOVA shows is mechanistically mediated by Zipfian word-length and subword compactness ($\SwadeshNounMeanFert{}$ vs.\ $\NonSwadeshNounMeanFert{}$ tokens/word; $t = \NounFertAncovaT{}$, $p = \NounFertAncovaP{}$).

Sections~\ref{sec:results-phylo} and~\ref{sec:results-colex} evaluate the encoder's geometry against independent typological benchmarks from historical linguistics and cross-linguistic lexical semantics.
In Section~\ref{sec:results-phylo}, we show that pairwise language distances in NLLB-200's encoder correlate strongly with Automated Similarity Judgment Program (ASJP) phonetic distances \citep{wichmann2022,jaeger2018} within language subfamilies ($\rho_s = \MantelWithinSubfamilyRho{}$, within-subfamily block-permutation $p = \MantelWithinSubfamilyBlockP{}$, dyadic $p = \MantelWithinSubfamilyP{}$, $n = \MantelWithinSubfamilyPairs{}$ pairs), including across within-subfamily language pairs written in completely disjoint scripts ($\rho_s = \MantelCrossScriptSubfamilyRho{}$ in the carrier frame; $\rho_s = \MantelBareCrossScriptSubfamilyRho{}$ on bare words, $n = \MantelCrossScriptSubfamilyPairs{}$).
Across unrelated language families ($n = \MantelBetweenFamilyPairs{}$ pairs, $\MantelBetweenFamilyPct{}\%$ of all pairs), ASJP lexical distance saturates near ceiling ($\rho_s = \MantelBetweenFamilyRho{}$), yielding a modest pooled Mantel correlation across all $\MantelNumLangs{}$ languages in the carrier frame ($\rho_s = \MantelRho{}$, $p = \MantelP{}$) that nearly doubles on native bare-word embeddings ($\rho_s = \MantelBareRho{}$, $p = \MantelBareP{}$).
In Section~\ref{sec:results-colex}, evaluating all $\binom{\NumConcepts}{2} = \ColexNumPairs{}$ Swadesh concept pairs against the CLICS\textsuperscript{3} cross-linguistic colexification database \citep{list2018,rzymski2020} reveals that concept pairs colexified across $\ge 3$ language families ($n = \ColexColCount{}$) exhibit substantially higher embedding similarity than never-colexified pairs ($n = \ColexNonColCount{}$; $U = \ColexU{}$, $p = \ColexP{}$, Cohen's $d = \ColexCohenD{}$), with a continuous positive rank correlation across all \ColexNumPairs{} pairs ($\rho_s = \ColexSpearmanRho{}$, $p = \ColexSpearmanP{}$).
This effect remains virtually unchanged when excluding identical word forms and cognates ($d = \ColexStringDistinctCohenD{}$), strengthens within semantic categories ($\rho_s = \ColexWithinCatRho{}$, $p = \ColexWithinCatRhoP{}$, $d = \ColexWithinCatCohenD{}$) and within parts of speech ($d = \ColexSamePOSCohenD{}$), and is even stronger on decontextualized bare words ($d = \ColexBareStringDistinctCohenD{}$, $p = \ColexBareStringDistinctP{}$).

Finally, Section~\ref{sec:results-domains} probes domain-specific conceptual subspaces and second-order relational offsets.
Within the color domain (\ColorNumColors{} basic terms across \ColorNumLanguages{} languages), the first principal component ($\ColorPConeVar{}\%$ of variance) directly recovers the \citet{berlin1969} evolutionary hierarchy from Stage~I (\emph{white, black}) through Stages~II--VI to Stage~VII derived terms (\emph{purple, pink, orange}; Spearman $\rho_s = \ColorBKRho{}$, $p = \ColorBKP{}$; Pearson $r = \ColorBKR{}$, $p = \ColorBKRp{}$), and replicates on decontextualized native bare words ($\rho_s = \ColorBareBKRho{}$, $p = \ColorBareBKP{}$) and after controlling for cross-lingual word length (partial $r = \ColorPartialLengthR{}$, $p = \ColorPartialLengthP{}$).
Across relational difference vectors $\mathbf{e}_{b,\ell} - \mathbf{e}_{a,\ell}$, while raw cosine consistency is mechanically coupled to baseline centroid separation $\|\hat{\boldsymbol{\mu}}_b - \hat{\boldsymbol{\mu}}_a\|_2$ ($r = \OffsetNormCorrR{}$, $p = \OffsetNormCorrP{}$), linear and concave distance-controlled residuals ($U = \OffsetResidU{}$, $p = \OffsetResidP{}$, $d = \OffsetResidCohenD{}$), doubly-centered perturbation coupling ($+\OffsetCouplingSemanticMean{}$ vs.\ $\OffsetCouplingControlMean{}$, $U = \OffsetCouplingU{}$, $p = \OffsetCouplingP{}$), and distance/POS-matched negative controls ($p = \OffsetMatchedResidP{}$) confirm genuine cross-lingual relational alignment.

All experiments, pre-computed data artifacts, and interactive 3D visualizations are publicly available via the open-source \textsc{InterpretCognates} platform at \url{https://github.com/kylemath/InterpretCognates}, with an interactive companion at \url{https://kylemath.github.io/InterpretCognates/}.

\section{Background}

Our work draws on two largely separate literatures: the geometry of multilingual neural representations and the cognitive science of how multilinguals organize meaning.
We review each in turn, highlighting the specific hypotheses that motivate our experiments.

\subsection{Multilingual Representation Geometry}

A central question in multilingual NLP is whether shared Transformer encoder models \citep{vaswani2017} learn language-neutral representations or merely co-locate language-specific subspaces.
\citet{pires2019} provided early evidence for the former, demonstrating that multilingual BERT \citep{devlin2019} supports zero-shot cross-lingual transfer on NER and POS tagging even between typologically distant languages, suggesting the emergence of shared syntactic abstractions \citep{hewitt2019}.
Subsequent work has refined this picture considerably.

\citet{chang2022} decomposed the representation space of XLM-R \citep{conneau2020} into language-sensitive and language-neutral subspaces via singular value decomposition of language- and token-level means.
They found that projecting out the low-dimensional language-sensitive subspace improves cross-lingual alignment on semantic tasks, implying that language identity occupies a compact shift subspace largely orthogonal to a higher-dimensional conceptual manifold.
Their geometric decomposition directly motivates our conceptual store experiment, which isolates the language-neutral component by subtracting per-language centroids and quantifies the effective dimensionality of each subspace.

The geometry of multilingual encoders is further complicated by anisotropy---the tendency of learned representations to cluster in a narrow cone rather than occupying the full available volume \citep{ethayarajh2019,rajaee2022}.
\citet{mu2018} proposed All-but-the-Top (ABTT), a post-processing method that removes the global mean and top-$k$ principal components from embeddings to restore isotropy.
Per-language mean-centering and global ABTT address complementary facets of anisotropy: per-language centering removes language-specific translation shifts $\boldsymbol{\ell}$, while ABTT removes dominant global variance directions across the combined corpus.

At a finer grain, \citet{voita2019,voita2021} demonstrated that individual attention heads and layers in Transformer translation models specialize for distinct linguistic functions across source and target processing, including positional, syntactic, and rare-token tracking.
\citet{foroutan2022} extended this line of work to the multilingual case, identifying language-neutral sub-networks within multilingual Transformers that activate consistently across languages for equivalent inputs.
These findings suggest that universality is not merely a global property of the representation space but is also reflected in modular internal structure.

Taken together, this literature establishes that multilingual Transformers encode both language-specific and language-neutral information in geometrically separable subspaces.
Our experiments test whether this geometric separation extends to NLLB-200 \citep{nllbteam2022,goyal2022}---a model trained explicitly for translation across 200 languages---and whether the language-neutral component exhibits structure predicted by cognitive science.

\subsection{Cognitive Science of Multilingual Representation}

The question of whether bilinguals and multilinguals maintain a shared conceptual store has been debated for decades.
The Revised Hierarchical Model \citep{kroll1994,kroll2010} posits that bilinguals access a common conceptual store through language-specific lexical representations, with direct concept--word links strengthening as a function of language exposure and proficiency—predicting tighter conceptual coupling for high-exposure than low-exposure languages.
The BIA+ model \citep{dijkstra2002} further proposes that bilingual word recognition involves an integrated lexical identification system interfacing with a shared semantic level and a task-decision gate.

Neuroimaging studies distinguish between form-level language processing and downstream conceptual hubs.
\citet{malikmoraleda2022} demonstrated across 45 languages from 12 families that the core fronto-temporal language network maintains consistent functional selectivity for lexical-syntactic processing across typologically diverse languages, while at the sub-lexical level \citet{thierry2007} showed via event-related potentials that bilinguals unconsciously activate native-language forms during second-language reading.
Downstream of form processing, \citet{correia2014} used searchlight multivariate decoding on fMRI data to show that the left anterior temporal lobe (ATL) supports cross-language classification of spoken animal concepts in Dutch--English bilinguals, consistent with an amodal conceptual hub.
More recently, \citet{deniz2025} used voxelwise encoding models on naturalistic narrative fMRI data in Chinese--English bilinguals to show that cortical semantic representations share a common cross-lingual core modulated both by global language-indicator shifts and by semantic-cluster-specific tuning rotations—paralleling both the global language shift $\boldsymbol{\ell}$ and the structured concept-by-language interaction residual $\boldsymbol{\epsilon}_{c,\ell}$ in our additive model.

Cross-linguistic universals from historical linguistics and typology provide independent external ground truth for evaluating these representations.
\citet{swadesh1952} identified a core vocabulary of basic concepts (body parts, kinship terms, natural phenomena, basic actions) that resists borrowing and evolves slowly across human languages.
The Automated Similarity Judgment Program (ASJP) database \citep{wichmann2022,jaeger2018} quantifies lexical phonetic distance across world languages using 40-item Swadesh wordlists in standardized phonetic transcription, providing a genealogical benchmark for our phylogenetic analysis.
In lexical typology, \citet{berlin1969} demonstrated that languages partition color space along an implicational evolutionary hierarchy from Stage~I (\emph{black, white}) and Stage~II (\emph{red}) through Stages~III--VI (\emph{green, yellow, blue, brown}) to Stage~VII (\emph{purple, pink, orange, grey}), while the CLICS\textsuperscript{3} database \citep{list2018,rzymski2020} catalogs cross-linguistic colexifications---concept pairs expressed by a single word form within a language, complementing monolingual lexical ontologies such as WordNet \citep{miller1995}---across thousands of languages.
Finally, \citet{vulic2020} established in \emph{Multi-SimLex} that cross-lingual lexical semantic alignment varies systematically by part of speech and concreteness (with concrete nouns aligning more tightly than adjectives and verbs), while \citet{mikolov2013} showed that linear relational offsets rely on approximate relational isomorphism across languages—hypotheses we test directly via category-stratified convergence and distance-controlled semantic offset vectors.

\section{Methods}
\label{sec:methods}

\subsection{Model, Lexical Corpora, and Extraction Regimes}

We probe the shared encoder of Meta's 600M-parameter distilled NLLB-200 translation model (\texttt{facebook/nllb-200-distilled-600M}; \citealp{nllbteam2022}), which consists of $L = \LayerwiseNumLayers{}$ Transformer encoder blocks with hidden dimension $d = 1024$ and a shared 256,000-token SentencePiece vocabulary across 200 languages.
As our primary lexical probe, we use the \NumConcepts{}-concept Swadesh core vocabulary \citep{swadesh1952} across \NumLanguages{} languages after excluding six languages (\texttt{ary\_Arab}, \texttt{bod\_Tibt}, \texttt{uzb\_Latn}, \texttt{knc\_Latn}, \texttt{san\_Deva}, \texttt{ban\_Latn}) whose tokenizations or norms produce degenerate PCA outliers ($>3\times\text{IQR}$ fence).
For targeted comparisons, we also evaluate two 60-concept non-Swadesh sets across a 39-language typologically balanced subset—a loanword-heavy modern set (e.g., \emph{coffee, radio, internet}) and a controlled concrete-noun set (e.g., \emph{hammer, basket, axe, spider})—as well as the \ColorNumColors{} basic color terms of \citet{berlin1969} across \ColorNumLanguages{} languages.

\begin{figure*}[htbp]
    \centering
    \includegraphics[width=\textwidth,height=0.82\textheight,keepaspectratio]{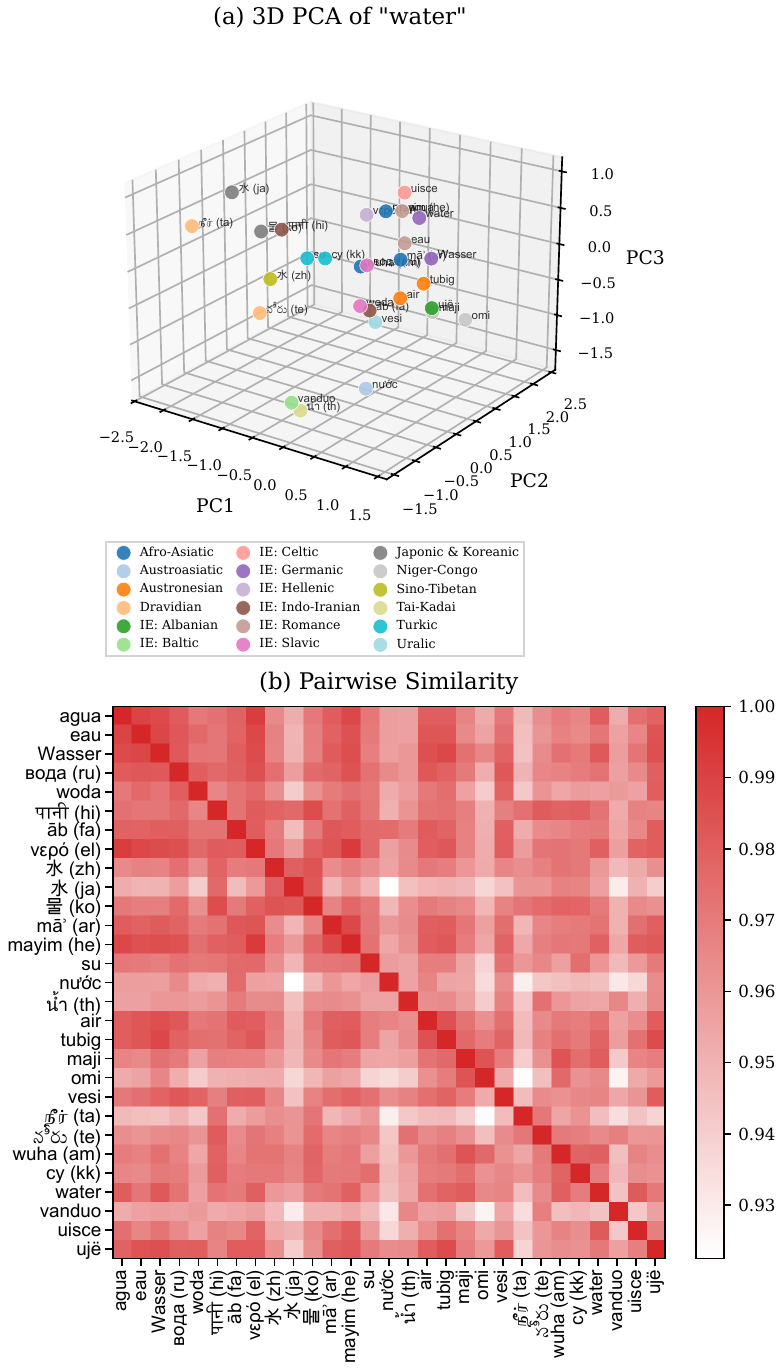}
    \caption{Embedding geometry for the concept ``water'' across 29 typologically diverse languages (embedded in ``\texttt{I drink }\emph{word}\texttt{.}'' ). (a) 3D PCA projection colored by language family shows genealogy-aligned subclusters within a shared neighborhood despite script and phonological diversity (RTL scripts shown in romanized transcription for display clarity). (b) Pairwise cosine similarity heatmap confirms both intra-family clustering (e.g., Romance, Slavic) and high cross-family similarity ($>0.93$).}
    \label{fig:water_manifold}
\end{figure*}

To separate intrinsic lexical geometry from contextual framing effects, we extract encoder representations under two complementary regimes.
In the \textbf{shared carrier-frame regime}, each target-language word $w_{c,\ell}$ is spliced into a fixed English carrier template (\texttt{"I saw a }$\{$\texttt{word}$\}$\texttt{ near the river."}, or \texttt{"I drink }$\{$\texttt{word}$\}$\texttt{."} for the Figure~\ref{fig:water_manifold} illustration) while setting the SentencePiece source-language tag to the target language (\texttt{tokenizer.src\_lang} $= \ell$). We locate the exact subword token span $T(w_{c,\ell})$ matching $w_{c,\ell}$ via stripped character-sequence alignment and mean-pool only those target-word hidden states at the final encoder block (or across all 12 Transformer encoder blocks, indexed as Layers~$0, \dots, 11$ corresponding to \texttt{hidden\_states[1:13]}). Across all $13{,}635$ $(\text{concept}, \text{language})$ pairs in our Swadesh corpus, exact target-word subword span isolation succeeds on $100.0\%$ of items ($13{,}635 / 13{,}635$, with zero fallbacks to full-sentence pooling), ensuring that English carrier tokens are never averaged into the target representation. Because self-attention mixes context information into the target token positions across layers, this regime provides a uniform grammatical slot that partially aligns cross-lingual centroids.
In the \textbf{decontextualized bare-word regime}, each native word $w_{c,\ell}$ is encoded in complete isolation with \texttt{tokenizer.src\_lang} $= \ell$ and zero English context tokens, mean-pooling its subword hidden states. Comparing the two regimes isolates what the model encodes about native word forms prior to contextual homogenization.

\subsection{Geometric Corrections and Evaluation Metrics}

Let $\mathbf{e}_{c,\ell} \in \mathbb{R}^d$ denote the extracted encoder representation for concept $c \in \{1, \dots, M\}$ in language $\ell \in \{1, \dots, N\}$, and let $\cos(\mathbf{u}, \mathbf{v}) = \frac{\mathbf{u}^\top \mathbf{v}}{\|\mathbf{u}\|_2 \|\mathbf{v}\|_2}$.

\paragraph{All-But-The-Top (ABTT) Isotropy Correction.}
Because raw Transformer representations occupy a narrow anisotropic cone \citep{ethayarajh2019,rajaee2022}, raw cosine similarities are systematically inflated.
Following \citet{mu2018}, we center the pooled $MN \times d$ corpus matrix by subtracting the global mean $\bar{\mathbf{e}} = \frac{1}{MN}\sum_{c,\ell}\mathbf{e}_{c,\ell}$ and project out the top $k$ principal components $\mathbf{u}_1, \dots, \mathbf{u}_k$ fitted across all $MN$ embeddings (with $k = 3$ as our reference and $k \in \{0, 1, 3, 5, 10\}$ for sensitivity analysis):
\begin{equation}
\tilde{\mathbf{e}}_{c,\ell} \;=\; (\mathbf{e}_{c,\ell} - \bar{\mathbf{e}}) \;-\; \sum_{j=1}^{k} \big((\mathbf{e}_{c,\ell} - \bar{\mathbf{e}})^\top \mathbf{u}_j\big)\mathbf{u}_j.
\label{eq:abtt}
\end{equation}

\paragraph{Cross-Lingual Concept Convergence.}
For each concept $c$, we define its cross-lingual convergence score $C(c)$ as the mean pairwise cosine similarity across all $\binom{N}{2}$ language pairs:
\begin{equation}
C(c) \;=\; \frac{2}{N(N-1)} \sum_{1 \le \ell < \ell' \le N} \cos(\tilde{\mathbf{e}}_{c,\ell},\, \tilde{\mathbf{e}}_{c,\ell'}).
\label{eq:convergence}
\end{equation}

\paragraph{Two-Way Additive Decomposition, Per-Language Mean-Centering, and the Conceptual Store Metric (CSM).}
In the balanced $M \times N$ design (Equation~\ref{eq:intro-decomposition}), let $\bar{\mathbf{e}} = \frac{1}{MN}\sum_{c,\ell}\mathbf{e}_{c,\ell}$ denote the grand mean, $\bar{\mathbf{e}}_c = \frac{1}{N}\sum_{\ell=1}^{N}\mathbf{e}_{c,\ell}$ the raw concept centroid, and $\bar{\mathbf{e}}_\ell = \frac{1}{M}\sum_{c=1}^{M}\mathbf{e}_{c,\ell}$ the raw language centroid.
Imposing standard zero-sum identifiability constraints ($\sum_{c=1}^{M}\hat{\boldsymbol{\mu}}_c = \mathbf{0}$ and $\sum_{\ell=1}^{N}\hat{\boldsymbol{\ell}} = \mathbf{0}$) yields orthogonal least-squares estimators $\hat{\boldsymbol{\mu}}_c = \bar{\mathbf{e}}_c - \bar{\mathbf{e}}$, $\hat{\boldsymbol{\ell}} = \bar{\mathbf{e}}_\ell - \bar{\mathbf{e}}$, and residual $\hat{\boldsymbol{\epsilon}}_{c,\ell} = \mathbf{e}_{c,\ell} - \bar{\mathbf{e}} - \hat{\boldsymbol{\mu}}_c - \hat{\boldsymbol{\ell}}$, partitioning the total centered sum of squares into exact additive components:
\begin{equation}
\sum_{c,\ell}\|\mathbf{e}_{c,\ell} - \bar{\mathbf{e}}\|_2^2 \;=\; N\sum_{c=1}^{M}\|\hat{\boldsymbol{\mu}}_c\|_2^2 \;+\; M\sum_{\ell=1}^{N}\|\hat{\boldsymbol{\ell}}\|_2^2 \;+\; \sum_{c,\ell}\|\hat{\boldsymbol{\epsilon}}_{c,\ell}\|_2^2.
\label{eq:ss-partition}
\end{equation}
Subtracting each language's mean vector $\bar{\mathbf{e}}_\ell$ yields language-centered representations $\mathbf{e}^{\text{cen}}_{c,\ell} = \mathbf{e}_{c,\ell} - \bar{\mathbf{e}}_\ell = \hat{\boldsymbol{\mu}}_c + \hat{\boldsymbol{\epsilon}}_{c,\ell}$ whose concept centroids are $\hat{\boldsymbol{\mu}}_c$.
Using cosine distance $d_{\cos}(\mathbf{u}, \mathbf{v}) = 1 - \cos(\mathbf{u}, \mathbf{v})$ on unit-normalized embeddings and denoting the resulting active concept centroids by $\hat{\boldsymbol{\mu}}_c$, we define the Conceptual Store Metric (CSM) as the ratio of mean between-concept centroid distance to mean within-concept cross-lingual distance:
\begin{equation}
\text{CSM} \;=\; \frac{\frac{2}{M(M-1)}\sum_{c < c'} d_{\cos}(\hat{\boldsymbol{\mu}}_c,\, \hat{\boldsymbol{\mu}}_{c'})}{\frac{1}{M}\sum_{c=1}^{M} \frac{2}{N(N-1)}\sum_{\ell < \ell'} d_{\cos}(\mathbf{e}_{c,\ell},\, \mathbf{e}_{c,\ell'})},
\label{eq:csm}
\end{equation}
with $95\%$ confidence intervals estimated via 1,000 non-self concept-level bootstrap resamples.
To rule out artifacts of global origin shifting or shared script, we benchmark language-centered CSM against a concept-permutation null distribution that independently permutes concept labels within each language, against global mean-centering ($\mathbf{e}_{c,\ell} - \bar{\mathbf{e}}$), and against a strict cross-script CSM computed exclusively over language pairs with disjoint writing systems ($\text{script}(\ell) \neq \text{script}(\ell')$).

\paragraph{External Typological Benchmarks (ASJP and CLICS\textsuperscript{3}).}
To test genealogical structure, we compute pairwise language distances $D_{\text{emb}}(\ell, \ell') = \frac{1}{M}\sum_{c=1}^{M} d_{\cos}(\mathbf{e}_{c,\ell}, \mathbf{e}_{c,\ell'})$ and compare them against lexical phonetic distances from ASJP v20 \citep{wichmann2022,jaeger2018}—computed as the mean Normalized Levenshtein Distance ($\text{NLD}(a,b) = \text{lev}(a,b)/\max(|a|,|b|)$, taking the minimum distance across documented transcription variants per concept; dividing by off-diagonal non-synonymous NLD to form \citeauthor{jaeger2018}'s \citeyear{jaeger2018} LDND metric yields virtually identical within-subfamily [$\rho_s = 0.983$ with NLD; $\rho_s = \MantelLDNDSubfamilyRho{}$ with embeddings] and global [$\rho_s = \MantelLDNDRho{}$] correlations) across shared 40-item Swadesh transcriptions for $n = \MantelNumLangs{}$ mapped languages—using a Spearman Mantel test with \MantelPermutations{} permutations globally, a within-subfamily block-permutation Mantel test ($9{,}999$ permutations permuting languages strictly within each genealogical subfamily), and cross-script and partial Mantel controls adjusting for shared writing system.
To test conceptual proximity against cross-linguistic polysemy, we link all \NumConcepts{} Swadesh concepts to standardized Concepticon glosses in CLICS\textsuperscript{3} \citep{list2018,rzymski2020} (unioning closely related action and pronoun sub-glosses such as \texttt{THOU}/\texttt{YOU} and \texttt{LIE DOWN}/\texttt{LIE [REST]}), counting the number of distinct language families attesting each of the $\binom{\NumConcepts}{2} = \ColexNumPairs{}$ concept pairs.
For each concept pair $(a,b)$, we evaluate both the mean within-language cosine similarity $\bar{s}_{\text{within}}(a,b) = \frac{1}{N}\sum_{\ell=1}^{N}\cos(\mathbf{e}_{a,\ell}, \mathbf{e}_{b,\ell})$ (which compares $a$ and $b$ inside each language's own frame) and the cross-lingual centroid similarity $\cos(\hat{\boldsymbol{\mu}}_a, \hat{\boldsymbol{\mu}}_b)$, comparing frequently colexified pairs ($\ge 3$ families, $n = \ColexColCount{}$) against never-colexified pairs ($0$ families, $n = \ColexNonColCount{}$) via a one-sided Mann-Whitney $U$ test, sample-size-weighted Cohen's $d$, and continuous Spearman rank correlation across string-distinct ($w_{a,\ell} \neq w_{b,\ell}$), cognate-excluded ($\text{NLD} \ge 0.2$), within-category, same-POS, and bare-word controls.

\paragraph{Semantic Offset Consistency and Distance-Controlled Residuals.}
For a concept pair $(a, b)$, let $\mathbf{d}_{\ell}(a,b) = \mathbf{e}_{b,\ell} - \mathbf{e}_{a,\ell}$ denote its offset vector in language $\ell$ and $\bar{\mathbf{d}}(a,b) = \frac{1}{N}\sum_{\ell=1}^{N}\mathbf{d}_{\ell}(a,b) = \hat{\boldsymbol{\mu}}_b - \hat{\boldsymbol{\mu}}_a$ denote the centroid offset.
Raw cross-lingual offset consistency is the mean cosine alignment with the centroid offset:
\begin{equation}
S(a, b) \;=\; \frac{1}{N}\sum_{\ell=1}^{N} \cos\!\big(\mathbf{d}_{\ell}(a,b),\, \bar{\mathbf{d}}(a,b)\big).
\label{eq:offset}
\end{equation}
Because $\mathbf{d}_{\ell}(a,b) = (\hat{\boldsymbol{\mu}}_b - \hat{\boldsymbol{\mu}}_a) + (\hat{\boldsymbol{\epsilon}}_{b,\ell} - \hat{\boldsymbol{\epsilon}}_{a,\ell})$ cancels the shared language vector $\hat{\boldsymbol{\ell}}$, $S(a,b)$ increases mechanically with baseline centroid separation $D(a,b) = \|\hat{\boldsymbol{\mu}}_b - \hat{\boldsymbol{\mu}}_a\|_2$ according to $\mathbb{E}[S(a,b)] \approx D / \sqrt{D^2 + \sigma_{\Delta\epsilon}^2}$, where $\sigma_{\Delta\epsilon}^2 = \sigma_{\epsilon_a}^2 + \sigma_{\epsilon_b}^2 - 2\sigma_{\epsilon_a}\sigma_{\epsilon_b}\cos(\hat{\boldsymbol{\epsilon}}_{a,\ell}, \hat{\boldsymbol{\epsilon}}_{b,\ell})$ decreases whenever language-specific perturbations of $a$ and $b$ co-rotate.
We therefore evaluate \OffsetSemanticCount{} semantically related pairs against both \OffsetControlCount{} cross-domain negative controls and \OffsetMatchedCount{} distance- and POS-matched never-colexified negative controls, testing linear and nonlinear concave ($D/\sqrt{D^2+c}$) distance-adjusted residuals $S_{\text{resid}}(a,b)$ (including leave-one-language-out centroid scoring) alongside the doubly-centered cross-concept perturbation coupling $\frac{1}{N}\sum_{\ell=1}^{N}\cos(\hat{\boldsymbol{\epsilon}}_{a,\ell}, \hat{\boldsymbol{\epsilon}}_{b,\ell})$ computed after subtracting both per-language shifts $\hat{\boldsymbol{\ell}}$ and concept centroids $\hat{\boldsymbol{\mu}}_c$.

\section{Results}
\label{sec:results}

We present our six core empirical analyses following the progression outlined in the Introduction: from the global decomposition of language and concept geometry across encoder layers (Section~\ref{sec:results-csm}), through the Swadesh convergence hierarchy and methodological controls (Sections~\ref{sec:results-hierarchy}--\ref{sec:results-confounds}), to external typological benchmarks (Sections~\ref{sec:results-phylo}--\ref{sec:results-colex}) and fine-grained domain geometry (Section~\ref{sec:results-domains}).
Table~\ref{tab:master-summary} summarizes the sample sizes, embedding regimes, test statistics, effect sizes, and primary conclusions across all six experiments; interactive 3D and 2D counterparts of every analysis, organized in matching section order, are available at \url{https://kylemath.github.io/InterpretCognates/}.

\begin{table*}[t]
\centering
\scriptsize
\setlength{\tabcolsep}{3.5pt}
\renewcommand{\arraystretch}{1.15}
\caption{Summary of primary experiments and control tests probing the shared encoder geometry of \texttt{facebook/nllb-200-distilled-600M}. All statistics are computed deterministically from model activations, linked via automated \LaTeX{} macros (\texttt{output/stats.tex}), and synchronized with the interactive browser companion.}
\label{tab:master-summary}
\vspace{0.2em}
\resizebox{\textwidth}{!}{%
\begin{tabular}{@{}>{\raggedright\arraybackslash}p{2.9cm}>{\raggedright\arraybackslash}p{2.5cm}>{\raggedright\arraybackslash}p{1.8cm}>{\raggedright\arraybackslash}p{2.5cm}>{\raggedright\arraybackslash}p{3.4cm}>{\raggedright\arraybackslash}p{3.8cm}@{}}
\toprule
\textbf{Experiment \& Control} & \textbf{Scope / Sample} & \textbf{Regime} & \textbf{Primary Metric} & \textbf{Statistic \& Effect Size} & \textbf{Principal Finding} \\
\midrule
1a.\ Conceptual Store (CSM) & \NumConcepts{} concepts $\times$ \NumLanguages{} langs & Bare word & Between/within distance ratio & $\ConceptualStoreBareRaw{} \to \ConceptualStoreBareCentered{}$ ($\ConceptualStoreBareImprovement{}\times$; null $\CSMPermNull{}$) & Language shift $\hat{\boldsymbol{\ell}}$ ($\CSMSSLangBare{}\%$ SS, $\text{PR}{=}\PRLangBare{}$) removed; cross-script $\CSMCrossScriptBareImprovement{}\times$ \\
1b.\ Conceptual Store (CSM) & \NumConcepts{} concepts $\times$ \NumLanguages{} langs & Carrier frame & Between/within distance ratio & $\ConceptualStoreRaw{} \to \ConceptualStoreCentered{}$ ($\ConceptualStoreImprovement{}\times$; null $\CSMPermNull{}$) & Frame reduces $\hat{\boldsymbol{\ell}}$ to $\CSMSSLangCtx{}\%$ SS; converges to same store ($\text{PR}{=}\PRConcept{}$) \\
1c.\ Layer-Wise Trajectory & \NumConcepts{} concepts $\times$ \LayerwiseNumLangs{} langs & Carrier (L0--L11) & Mean convergence \& CSM & L0: $\LayerwiseInputConv{} \to$ L\LayerwisePeakLayer{}: $\LayerwisePeakConv{}$ (L11: $\LayerwiseFinalConv{}$) & Peak $C(c)$ at L\LayerwisePeakLayer{}; CSM inflection at L\LayerwisePhaseTrans{} ($\LayerwiseFinalCSM{}$ at L11) \\
\addlinespace[2pt]
2a.\ Swadesh Hierarchy \& ANOVA & \NumConcepts{} concepts $\times$ \NumLanguages{} langs & Carrier (ABTT $k{=}3$) & Pairwise $C(c)$ \& 8-cat.\ ANOVA & $\eta^2 = \CatAnovaEtaPct{}\%$, $F(7,93) = \CatAnovaF{}$ ($p = \CatAnovaP{}$) & Category explains $20\times$ surface form; partial $F = \CatPartialFertF{}$ \\
2b.\ Isotropy Sensitivity & $k \in \{0, 1, 3, 5, 10\}$ & Carrier frame & Spearman rank corr.\ $\rho_s$ & $\rho_s = \IsotropySpearmanRho{}$ ($p = \IsotropySpearmanP{}$) & Concept ranking invariant to PC removal ($\rho_s \ge \IsotropyMinRho{}$) \\
2c.\ Carrier vs.\ Bare Word & \NumConcepts{} concepts $\times$ \NumLanguages{} langs & Both regimes & Rank $\rho_s$ \& paired $t$-test & $\rho_s = \CarrierBaselineRho{}$ ($p = \CarrierBaselineP{}$); $\Delta\mu = {+}\CarrierBaselineMeanDiff{}$ & Rank order preserved; frame shifts baseline ($t = \CarrierBaselineTstat{}$) \\
\addlinespace[2pt]
3a.\ Surface \& Fertility Controls & \NumConcepts{} concepts $\times$ \DecompNumLatinLangs{} Latin langs & Carrier (ABTT $k{=}3$) & OLS regression $R^2$ & Ortho $R^2 = \DecompRsqOrtho{}$; Phon $R^2 = \DecompRsqPhon{}$; Fert $R^2 = \DecompRsqFert{}$ & Neither surface overlap nor fertility drives Swadesh 101 \\
3b.\ POS \& Strict Concrete Nouns & \PosNounSwadeshCount{} Swadesh vs.\ \ControlledSwadeshCompNumNonSwadesh{}/\StrictNounCount{} ctrl & Carrier (\LayerwiseNumLangs{} langs) & Mann-Whitney $U$ \& ANCOVA & ABTT $p = \PosNounCorrCompP{}$ ($d = \PosNounCorrCompCohenD{}$); strict $d = \StrictNounCorrCohenD{}$ & Mediated by Zipfian subword compactness ($p = \NounFertAncovaP{}$) \\
\addlinespace[2pt]
4a.\ Subfamily Phylogeny & \MantelWithinSubfamilyPairs{} within-subfamily pairs & Carrier \& Bare & Block-perm.\ \& cross-script $\rho_s$ & $\rho_s = \MantelWithinSubfamilyRho{}$ (LDND $\MantelLDNDSubfamilyRho{}$, block $p = \MantelWithinSubfamilyBlockP{}$) & Genealogical signal holds across disjoint scripts ($n = \MantelCrossScriptSubfamilyPairs{}$) \\
4b.\ Global Mantel Test & $\binom{\MantelNumLangs}{2} = \MantelTotalPairs{}$ lang pairs & Carrier vs.\ Bare & Mantel $\rho_s$ (\MantelPermutations{} perms) & Carrier $\rho_s = \MantelRho{}$ ($p = \MantelP{}$); Bare $\rho_s = \MantelBareRho{}$ ($p = \MantelBareP{}$) & Cross-family ASJP saturation ($\MantelBetweenFamilyPct{}\%$ pairs) attenuates global $\rho_s$ \\
\addlinespace[2pt]
5.\ CLICS\textsuperscript{3} Colexification & $\binom{\NumConcepts}{2} = \ColexNumPairs{}$ concept pairs & Carrier \& Bare & String-distinct \& within-cat.\ $U, \rho_s$ & $U = \ColexU{}$, $p = \ColexP{}$, $d = \ColexCohenD{}$; within-cat.\ $\rho_s = \ColexWithinCatRho{}$ & Holds string-distinct ($d = \ColexStringDistinctCohenD{}$) \& on bare words ($d = \ColexBareStringDistinctCohenD{}$) \\
\addlinespace[2pt]
6a.\ Berlin \& Kay Colors & \ColorNumColors{} color terms $\times$ \ColorNumLanguages{} langs & Carrier \& Bare & PC1 vs.\ B\&K evolutionary stage & Carrier $\rho_s = \ColorBKRho{}$ ($p = \ColorBKP{}$); Bare $\rho_s = \ColorBareBKRho{}$ ($p = \ColorBareBKP{}$) & PC1 recovers Stage I $\to$ VII; length-controlled $r = \ColorPartialLengthR{}$ ($p = \ColorPartialLengthP{}$) \\
6b.\ Relational Offsets & \OffsetSemanticCount{} semantic vs.\ \OffsetControlCount{}/\OffsetMatchedCount{} ctrl & Carrier \& Bare & Residual $S_{\text{resid}}$ \& coupling & $U = \OffsetResidU{}$, $p = \OffsetResidP{}$, $d = \OffsetResidCohenD{}$; coupling $p = \OffsetCouplingP{}$ & Exceeds linear/concave \& POS-matched controls ($p = \OffsetMatchedResidP{}$) \\
\bottomrule
\end{tabular}%
}
\end{table*}

\subsection{Additive Language-Shift Geometry and Layer-Wise Emergence}
\label{sec:results-csm}

We first test our central structural hypothesis (Equation~\ref{eq:intro-decomposition}): that NLLB-200's encoder representations decompose into a shared conceptual centroid $\hat{\boldsymbol{\mu}}_c$ plus an additive per-language shift $\hat{\boldsymbol{\ell}}$.
Applying the exact two-way sum-of-squares partition (Equation~\ref{eq:ss-partition}) across all $101 \times 135 = 13{,}635$ embeddings reveals how contextual framing redistributes variance between language identity and concept identity.
In the decontextualized bare-word regime, concept centroids $\hat{\boldsymbol{\mu}}_c$ account for $\CSMSSConceptBare{}\%$ of total centered variance, per-language shifts $\hat{\boldsymbol{\ell}}$ account for $\CSMSSLangBare{}\%$, and the concept-by-language interaction residual $\hat{\boldsymbol{\epsilon}}_{c,\ell}$ accounts for $\CSMSSResidBare{}\%$.
In the shared carrier-frame regime (\texttt{"I saw a }$\{$\texttt{word}$\}$\texttt{ near the river."}), self-attention with the shared context tokens reduces the per-language shift variance nearly sevenfold to $\CSMSSLangCtx{}\%$, while concept centroids expand to $\CSMSSConceptCtx{}\%$ and the residual remains stable at $\CSMSSResidCtx{}\%$.
Furthermore, eigendecomposition of the doubly-centered distance matrices reveals a sharp dimensionality asymmetry generalizing \citet{chang2022}: the language-shift subspace $\{\hat{\boldsymbol{\ell}}\}$ is concentrated in a low-dimensional manifold with participation ratio $\text{PR} = (\sum_i \lambda_i)^2 / \sum_i \lambda_i^2 = \PRLangBare{}$ on bare words ($\PRLangBareTopTenPct{}\%$ of variance in the top 10 principal components; $\text{PR} = \PRLangCtx{}$ in the carrier frame), whereas the conceptual manifold $\{\hat{\boldsymbol{\mu}}_c\}$ spans nearly three times the effective dimensionality ($\text{PR} = \PRConcept{}$, with the top 10 components capturing only $\PRConceptTopTenPct{}\%$ of variance).

\begin{figure}[htbp]
  \centering
  \includegraphics[width=\columnwidth]{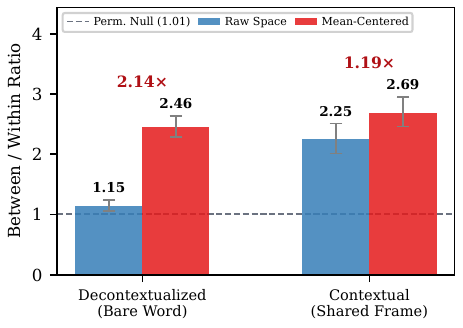}
  \caption{Conceptual Store Metric (CSM; ratio of mean between-concept centroid cosine distance to mean within-concept cross-lingual cosine distance) across \NumConcepts{} Swadesh concepts and \NumLanguages{} languages, comparing raw vs.\ language-centered embeddings under both extraction regimes against the concept-permutation null ($\text{CSM}_{\text{null}} = \CSMPermNull{}$, dashed grey line). Error bars denote 95\% bootstrap confidence intervals ($1{,}000$ non-self concept resamples). Decontextualized bare words exhibit strong raw language-identity offsets ($\text{CSM} = \ConceptualStoreBareRaw{}$) that are removed by per-language mean-centering ($\ConceptualStoreBareCentered{}$, a $\ConceptualStoreBareImprovement{}\times$ improvement), converging to nearly the same language-independent conceptual store as the shared carrier frame ($\ConceptualStoreRaw{} \to \ConceptualStoreCentered{}$, $\ConceptualStoreImprovement{}\times$).}
  \label{fig:conceptual_store}
\end{figure}

To evaluate how strongly representations cluster by concept rather than by language, we compute the Conceptual Store Metric (CSM; Equation~\ref{eq:csm}) before and after per-language mean-centering under both extraction regimes, alongside three null and confound controls (Figure~\ref{fig:conceptual_store}).

In the decontextualized bare-word regime, where native word forms are encoded with zero surrounding English context, the raw CSM is $\ConceptualStoreBareRaw{}$: between-concept distance only slightly exceeds the random concept-permutation null ($\text{CSM}_{\text{null}} = \CSMPermNull{} \pm 0.0004$, obtained by independently permuting concept labels within each language) because each language occupies a distinct subspace offset by $\hat{\boldsymbol{\ell}}$.
Subtracting each language's centroid $\hat{\boldsymbol{\ell}}$ increases the bare-word CSM to $\ConceptualStoreBareCentered{}$, a $\ConceptualStoreBareImprovement{}\times$ improvement ($+143\%$ above the permutation null, $p < 0.0001$; $95\%$ bootstrap CIs non-overlapping).
Crucially, this gain cannot be explained by simple global origin shifting (global mean-centering $\mathbf{e}_{c,\ell} - \bar{\mathbf{e}}$ yields a bare-word CSM of $\CSMBareGlobalCentered{}$ because per-language shifts $\hat{\boldsymbol{\ell}}$ continue to inflate within-concept cross-lingual dispersion at $0.574$, whereas per-language centering collapses within-concept dispersion to $0.411$).
In the shared carrier-frame regime, self-attention with the shared context tokens partially aligns the language centroids prior to post-processing, yielding a higher raw CSM of $\ConceptualStoreRaw{}$ that increases to $\ConceptualStoreCentered{}$ ($\ConceptualStoreImprovement{}\times$) after per-language mean-centering.
Moreover, when we restrict the CSM denominator exclusively to \textbf{strictly cross-script language pairs} ($\text{script}(\ell) \neq \text{script}(\ell')$, where zero orthographic or subword token overlap is possible), per-language mean-centering increases bare-word CSM by $\CSMCrossScriptBareImprovement{}\times$ (from $\CSMCrossScriptBareRaw{}$ to $\CSMCrossScriptBareCentered{}$) and carrier-frame CSM by $\CSMCrossScriptCtxImprovement{}\times$ (from $\CSMCrossScriptCtxRaw{}$ to $\CSMCrossScriptCtxCentered{}$).
Thus, across both extraction regimes and even across completely disjoint writing systems, NLLB-200 maps native lexical items onto a shared conceptual manifold obscured primarily by an additive language vector $\hat{\boldsymbol{\ell}}$.

\begin{figure*}[htbp]
  \centering
  \includegraphics[width=\textwidth]{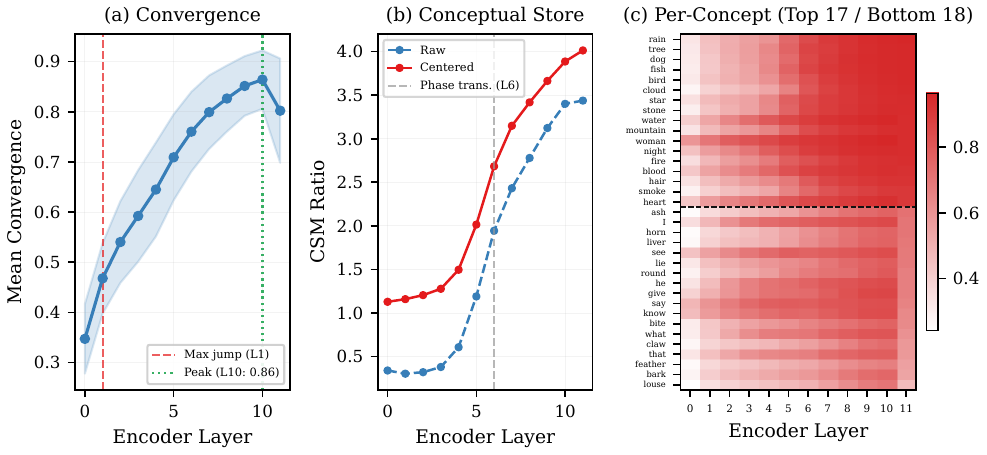}
  \caption{Layer-wise emergence of cross-lingual semantic structure across all \LayerwiseNumLayers{} encoder layers on the \LayerwiseNumLangs{}-language typologically balanced subset. (a) Mean Swadesh convergence rises from $\LayerwiseInputConv{}$ at the first encoder block (Layer~0), exhibits its steepest single-layer gain at Layer~\LayerwiseEmergenceLayer{}, reaches peak cross-lingual alignment at Layer~\LayerwisePeakLayer{} ($\mu = \LayerwisePeakConv{}$), and dips slightly at the final encoder layer (Layer~11, $\mu = \LayerwiseFinalConv{}$) as representations specialize for decoder cross-attention. (b) Raw and language-centered CSM across layers, with the steepest inflection in centered concept separability occurring at Layer~\LayerwisePhaseTrans{} and reaching $\LayerwiseFinalCSM{}$ at Layer~11. (c) Per-concept convergence heatmap across layers for the 17 highest- and 18 lowest-converging Swadesh concepts (sorted from highest to lowest final convergence), showing that concrete nouns and natural phenomena align several layers earlier than abstract or polysemous items.}
  \label{fig:layerwise}
\end{figure*}

To trace how this shared manifold is constructed across the depth of the network, we repeat both analyses at every encoder layer $l \in \{0, \dots, 11\}$ across our \LayerwiseNumLangs{}-language diverse subset (Figure~\ref{fig:layerwise}).
Cross-lingual convergence does not increase strictly monotonically to the final layer; rather, it rises sharply from $\LayerwiseInputConv{}$ at Layer~0 (largest single-step gain at Layer~\LayerwiseEmergenceLayer{}) to a \textbf{peak convergence of $\LayerwisePeakConv{}$ at Layer~\LayerwisePeakLayer{}}, before declining slightly to $\LayerwiseFinalConv{}$ at Layer~11 (Figure~\ref{fig:layerwise}a).
This late-layer specialization parallels findings in Transformer encoders \citep{voita2019,voita2021,tenney2019}, and decomposing the CSM numerator and denominator reveals its exact geometric mechanism.
Between Layer~\LayerwisePhaseTrans{} and Layer~\LayerwisePeakLayer{}, raw between-concept centroid cosine distance remains virtually constant ($\LayerSixRawBetween{} \to \LayerTenRawBetween{}$) while within-concept cross-lingual dispersion contracts by $\LayerWithinContractPct{}\%$ ($0.240 \to 0.136$), driving language-centered CSM to $\LayerTenCenteredCSM{}$.
From Layer~\LayerwisePeakLayer{} to Layer~11, within-concept distance expands by $+\LayerWithinExpandPct{}\%$ ($0.136 \to 0.198$, lowering mean $C(c)$ from $\LayerwisePeakConv{}$ to $\LayerwiseFinalConv{}$), \textbf{but between-concept centroid separation expands even faster by $+\LayerBetweenGainPct{}\%$ ($\LayerTenRawBetween{} \to \LayerElevenRawBetween{}$)} as the final encoder LayerNorm and feedforward block open the anisotropic cone for decoder cross-attention, lifting language-centered CSM from $\LayerTenCenteredCSM{}$ to $\LayerwiseFinalCSM{}$ (Figure~\ref{fig:layerwise}b).
Moreover, disaggregating the Layer~10 $\to$ Layer~11 transition by grammatical category (Figure~\ref{fig:layerwise}c) shows that concrete nouns fitting the carrier template's nominal slot (\emph{rain, tree, dog, fish, bird, cloud, star}) do not dip at all—peaking at Layer~11 ($C(c) = 0.95$--$0.96$)—whereas the late-layer divergence is driven selectively by grammatical Pronouns ($\Delta = \LayerPronounDip{}$) and Actions/verbs ($\Delta = \LayerActionDip{}$) unpacking target-language morphosyntactic features (person, number, valency) for the decoder.

\begin{figure*}[htbp]
  \centering
  \includegraphics[width=\textwidth]{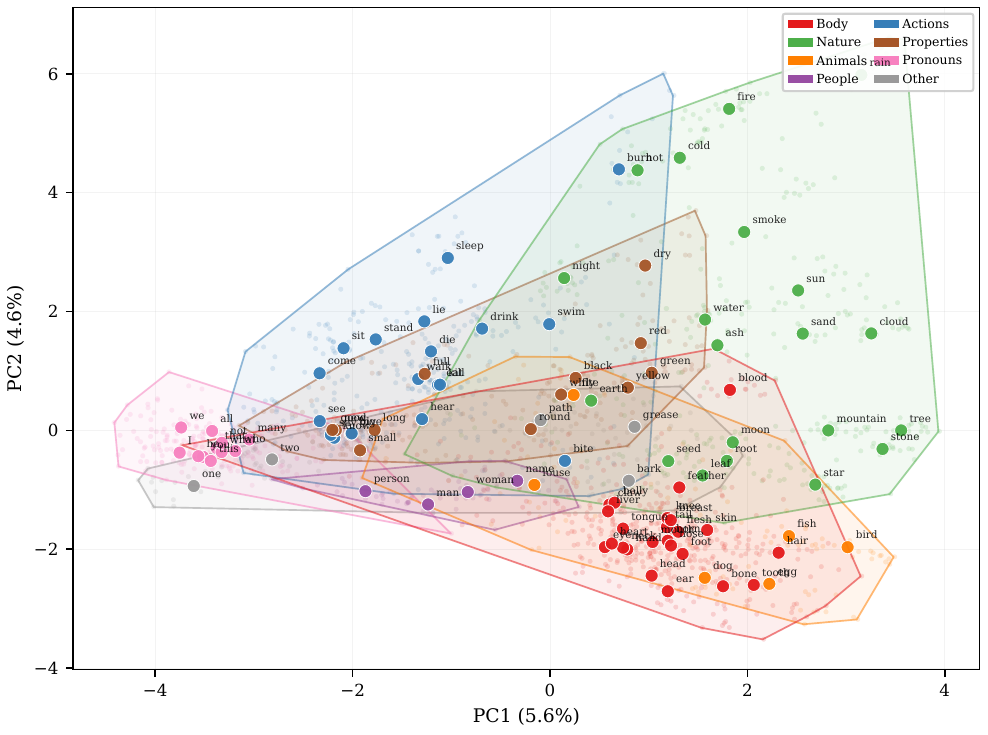}
  \caption{Two-dimensional PCA projection of Swadesh concept embeddings pooled across 19 language families (\NumConcepts{} concepts $\times$ \NumLanguages{} languages). Translucent dots represent per-family concept centroids; shaded convex hulls enclose semantic categories; large opaque markers indicate global cross-lingual centroids $\boldsymbol{\mu}_c$. Concept centroids organize by semantic domain (e.g., Pronouns, People, Body Parts, Nature) rather than collapsing by language family.}
  \label{fig:concept_map}
\end{figure*}

Projecting the resulting concept centroids $\boldsymbol{\mu}_c$ and per-family sub-centroids into a shared 2D PCA plane (Figure~\ref{fig:concept_map}) confirms visually that NLLB-200 organizes the core lexicon into coherent semantic neighborhoods—separating pronouns, human nouns, body parts, and natural phenomena across all 19 language families.

\subsection{Swadesh Convergence Hierarchy, Isotropy Robustness, and Carrier-Frame Baseline}
\label{sec:results-hierarchy}

Motivated by the single-concept manifold for ``water'' (Figure~\ref{fig:water_manifold}), we evaluate cross-lingual convergence $C(c)$ across all \NumConcepts{} Swadesh concepts and \NumLanguages{} languages after ABTT isotropy correction ($k = 3$).
Mean convergence across the full Swadesh inventory is $\SwadeshMean{} \pm \SwadeshStd{}$, spanning a wide dynamic range from $\SwadeshMin{}$ (\emph{\SwadeshBottomConcept{}}) to $\SwadeshMax{}$ (\emph{\SwadeshTopConcept{}}).

\begin{figure*}[htbp]
  \centering
  \includegraphics[width=\textwidth]{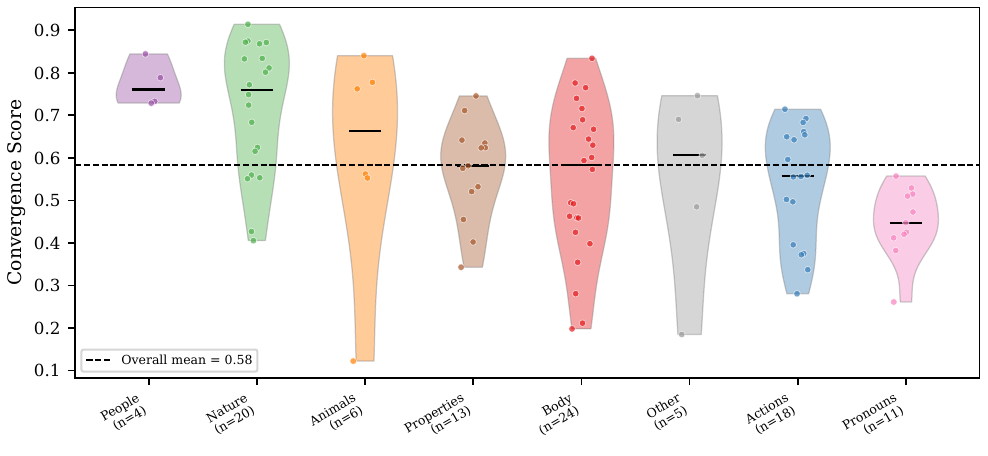}
  \caption{Cross-lingual convergence by semantic category across \NumConcepts{} Swadesh concepts (\NumLanguages{} languages, ABTT-corrected $k=3$). Violins and jittered points show per-concept scores; the vertical dashed line indicates the grand mean ($\SwadeshMean{}$). Perceptually grounded noun categories (People $\mu = \CatPeopleMean{}$, Nature $\mu = \CatNatureMean{}$) converge most strongly, whereas grammatical Pronouns ($\mu = \CatPronounsMean{}$) converge least.}
  \label{fig:category_summary}
\end{figure*}

\begin{figure}[htbp]
  \centering
  \includegraphics[width=\columnwidth]{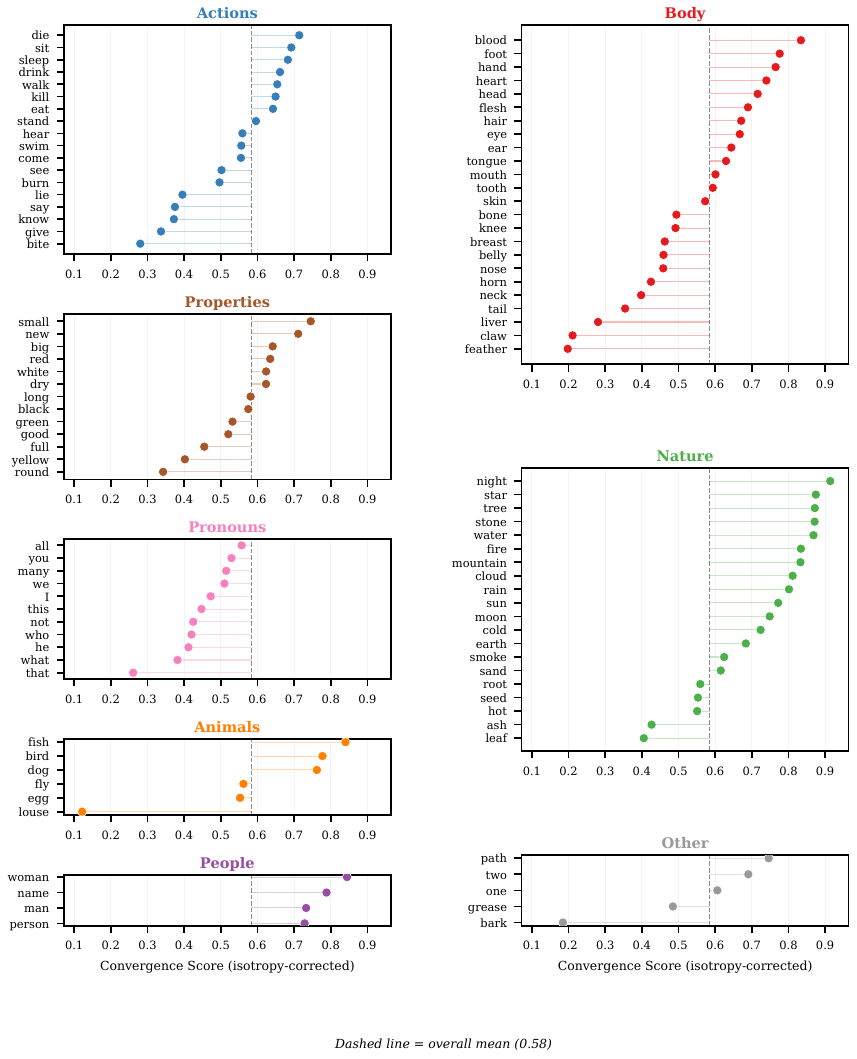}
  \caption{Per-concept ABTT-corrected convergence scores for all \NumConcepts{} Swadesh items grouped by semantic category (sorted by category mean). Individual within-category outliers reflect English lexical polysemy (e.g., \emph{bark, lie, fly}) or low-frequency rural/archaic vocabulary (\emph{louse, grease}).}
  \label{fig:category_detail}
\end{figure}

Grouping concepts into eight semantic categories exposes a clear, statistically robust hierarchy that mirrors the cross-lingual concreteness and part-of-speech gradient documented in \emph{Multi-SimLex} \citep{vulic2020} (Figure~\ref{fig:category_summary} and Figure~\ref{fig:category_detail}): People terms ($\CatPeopleMean{} \pm \CatPeopleStd{}$) and Nature terms ($\CatNatureMean{} \pm \CatNatureStd{}$) achieve the tightest cross-lingual alignment, followed by Animals ($\CatAnimalsMean{} \pm \CatAnimalsStd{}$), Properties ($\CatPropertiesMean{} \pm \CatPropertiesStd{}$), Body parts ($\CatBodyMean{} \pm \CatBodyStd{}$), Other ($\CatOtherMean{} \pm \CatOtherStd{}$), and Actions/verbs ($\CatActionsMean{} \pm \CatActionsStd{}$), while grammatical Pronouns/deictics rank lowest ($\CatPronounsMean{} \pm \CatPronounsStd{}$).
A one-way ANOVA across the eight semantic categories confirms that semantic domain accounts for \textbf{$\eta^2 = \CatAnovaEtaPct{}\%$ of the total variance in cross-lingual convergence} ($F(7, 93) = \CatAnovaF{}$, $p = \CatAnovaP{}$; non-parametric Kruskal-Wallis $H = \CatKWH{}$, $p = \CatKWP{}$)—more than twenty times the variance explained by surface orthography or phonology ($R^2 \le \DecompRsqOrtho{}$; Section~\ref{sec:results-confounds}).
This category effect replicates on decontextualized native bare words ($\eta^2 = \CatBareAnovaEtaPct{}\%$, $F(7, 93) = \CatBareAnovaF{}$, $p = \CatBareAnovaP{}$) and strengthens further in a partial $F$-test controlling simultaneously for cross-lingual SentencePiece subword fertility and character length ($F(7, 91) = \CatPartialFertF{}$, $p = \CatPartialFertP{}$).

Inspection of individual within-category outliers (Figure~\ref{fig:category_detail}) identifies two distinct sources of dispersion at the bottom of the ranking: language-specific grammaticalization in pronouns and demonstratives (\emph{we, this, that, who, what}), which encode clusivity, gender, honorifics, or spatial deixis differently across languages, and English lexical polysemy or low corpus frequency in items such as \emph{bark} (tree cortex vs.\ dog vocalization), \emph{lie} (recline vs.\ falsehood), \emph{fly} (insect vs.\ verb of motion), and \emph{louse} ($\SwadeshMin{}$, which is low-frequency in modern web-crawled parallel corpora).

\begin{figure*}[htbp]
  \centering
  \includegraphics[width=\textwidth]{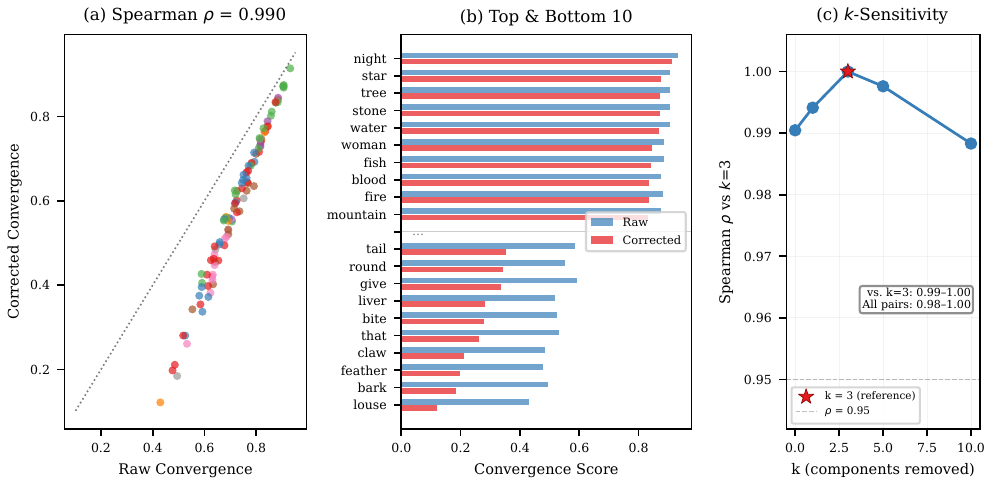}
  \caption{Validation of All-But-The-Top (ABTT) isotropy correction. (a) Raw vs.\ ABTT-corrected ($k=3$) convergence across all \NumConcepts{} concepts ($\rho_s = \IsotropySpearmanRho{}$, $p = \IsotropySpearmanP{}$). Removing the top 3 principal components eliminates the baseline anisotropic floor ($\sim 0.43$--$0.91 \to \SwadeshMin{}$--$\SwadeshMax{}$) while preserving rank order. (b) Top-10 and bottom-10 concepts before and after correction. (c) Hyperparameter sensitivity across $k \in \{0, 1, 3, 5, 10\}$: all rankings correlate at $\rho_s \in [\IsotropyMinRho{}, \IsotropyMaxRho{}]$ with the $k=3$ reference.}
  \label{fig:isotropy_validation}
\end{figure*}

Figure~\ref{fig:isotropy_validation} confirms that this hierarchy is not an artifact of ABTT preprocessing.
Raw and ABTT-corrected ($k=3$) concept rankings exhibit near-perfect rank agreement ($\rho_s = \IsotropySpearmanRho{}$, $p = \IsotropySpearmanP{}$; Figure~\ref{fig:isotropy_validation}a--b), and varying the number of removed principal components across $k \in \{0, 1, 3, 5, 10\}$ yields Spearman correlations spanning $\IsotropyKRange{}$ across all pairs (Figure~\ref{fig:isotropy_validation}c), with mean convergence exhibiting a sharp knee from $k = 0$ ($\IsotropyKZeroMean{}$) to $k = 1$ ($\IsotropyKOneMean{}$) before plateauing across $k = 3$ ($\IsotropyKThreeMean{}$) to $k = 10$ ($\IsotropyKTenMean{}$).

\begin{figure*}[!t]
  \centering
  \includegraphics[width=0.95\textwidth]{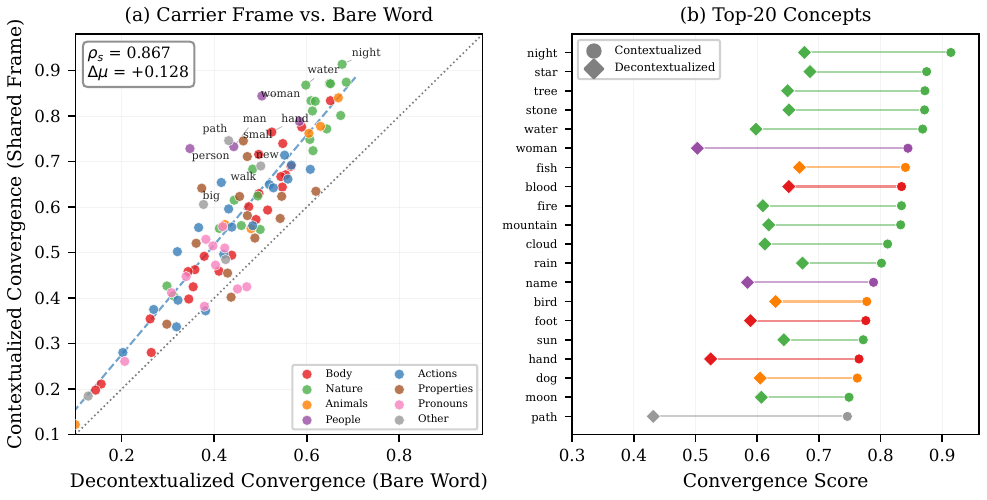}
  \caption{Comparison of Swadesh concept convergence in the shared English carrier frame vs.\ decontextualized native bare words across \NumLanguages{} languages. (a) Concept-level scatter shows strong rank-order invariance ($\rho_s = \CarrierBaselineRho{}$, $p = \CarrierBaselineP{}$) alongside a systematic upward shift in absolute convergence induced by the shared English carrier frame ($\mu_{\text{ctx}} = \CarrierCtxMean{}$ vs.\ $\mu_{\text{bare}} = \CarrierDectxMean{}$, $\Delta\mu = +\CarrierBaselineMeanDiff{}$, paired $t = \CarrierBaselineTstat{}$, $p = \CarrierBaselineTp{}$). (b) Slopegraph of top-ranked concepts confirming stability of concrete universals across both regimes.}
  \label{fig:carrier_baseline}
\end{figure*}

Similarly, comparing the shared carrier-frame regime against the decontextualized bare-word baseline across all \NumConcepts{} concepts (Figure~\ref{fig:carrier_baseline}) reveals two complementary effects.
First, the relative ordering of concepts is highly invariant across regimes ($\rho_s = \CarrierBaselineRho{}$, $p = \CarrierBaselineP{}$), confirming that concrete nouns converge more strongly than abstract or grammatical items even when embedded as isolated native words.
Second, as expected from self-attention over shared context tokens, the English carrier frame shifts absolute convergence upward by $\Delta\mu = +\CarrierBaselineMeanDiff{}$ (from $\mu_{\text{bare}} = \CarrierDectxMean{}$ to $\mu_{\text{ctx}} = \CarrierCtxMean{}$; paired $t = \CarrierBaselineTstat{}$, $p = \CarrierBaselineTp{}$).
Thus, the carrier frame acts as an additive baseline shift rather than generating the semantic hierarchy.

\subsection{Ruling Out Surface-Form Confounds and Isolating Zipfian Subword Compactness}
\label{sec:results-confounds}

A natural concern is that cross-lingual embedding similarity might simply reflect shared orthography, cognate roots, international loanwords, or subword tokenization length rather than conceptual alignment.
We evaluate these alternative explanations using three targeted tests.

\begin{figure*}[htbp]
  \centering
  \includegraphics[width=\textwidth]{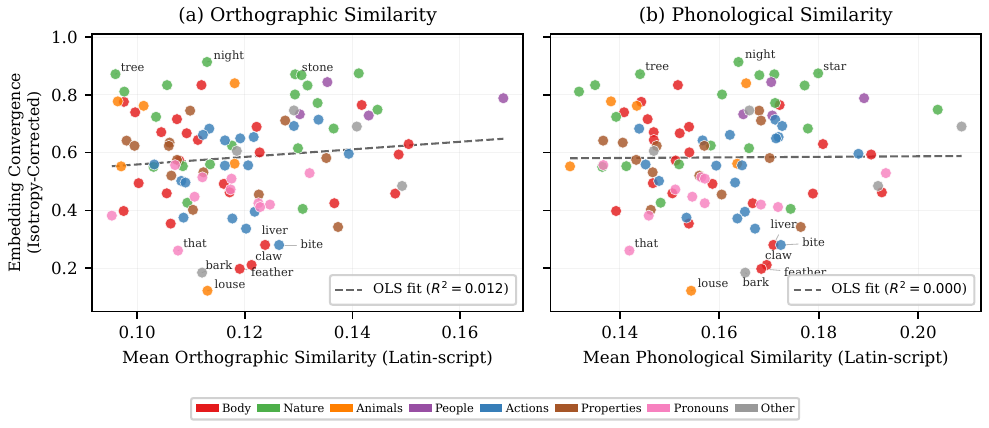}
  \caption{Swadesh convergence ($y$-axis, ABTT-corrected $k=3$) vs.\ cross-lingual surface-form similarity across all \DecompNumLatinLangs{} Latin-script languages in our sample. (a) Mean orthographic similarity ($1 - \text{NLD}$ on raw word forms) explains $R^2 = \DecompRsqOrtho{}$ ($p = \DecompPOrtho{}$). (b) Mean phonological similarity ($1 - \text{NLD}$ after voicing/diacritic normalization) explains $R^2 = \DecompRsqPhon{}$ ($p = \DecompPPhon{}$). Over $98.7\%$ of the variance in Swadesh convergence is independent of surface-form overlap.}
  \label{fig:swadesh}
\end{figure*}

First, across all \DecompNumLatinLangs{} Latin-script languages in our dataset (evaluated across all pairs without truncation), we regress ABTT-corrected concept convergence against mean cross-lingual orthographic similarity ($1 - \text{NLD}$), phonologically normalized similarity, character length, and SentencePiece subword fertility (tokens per word) across the \NumConcepts{} Swadesh concepts (Figure~\ref{fig:swadesh}).
Neither orthographic overlap ($R^2 = \DecompRsqOrtho{}$, slope $= \DecompSlopeOrtho{}$, $p = \DecompPOrtho{}$), phonological similarity ($R^2 = \DecompRsqPhon{}$, slope $= \DecompSlopePhon{}$, $p = \DecompPPhon{}$), character length ($R^2 = \DecompRsqLen{}$, $p = \DecompPLen{}$), nor subword fertility ($R^2 = \DecompRsqFert{}$, $p = \DecompPFert{}$) significantly predicts convergence across the Swadesh list (in fact, short function words and pronouns have the lowest subword fertility yet the lowest convergence).
Top-ranked Swadesh concepts such as \emph{night} ($\SwadeshMax{}$) and \emph{water} achieve near-maximal embedding alignment despite low cross-lingual orthographic similarity ($\sim \OrthoSimSwadesh{}$).

\begin{figure*}[htbp]
  \centering
  \includegraphics[width=\textwidth]{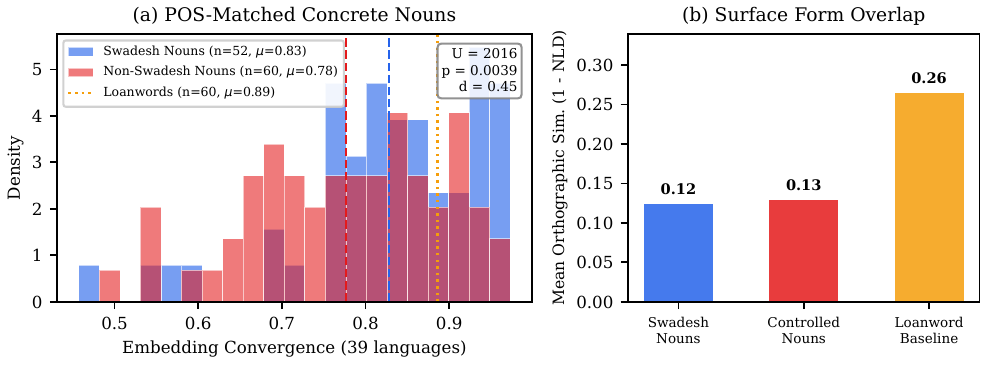}
  \caption{Controlling for loanword and part-of-speech confounds in Swadesh vs.\ non-Swadesh comparisons (\LayerwiseNumLangs{}-language subset). (a) When matched on part of speech and low orthographic overlap, Swadesh core concrete nouns ($n = \PosNounSwadeshCount{}$, $\mu = \PosNounSwadeshMean{}$) achieve significantly higher cross-lingual convergence than controlled non-Swadesh concrete nouns ($n = \ControlledSwadeshCompNumNonSwadesh{}$, $\mu = \PosNounNonSwadeshMean{}$; Mann-Whitney $U = \PosNounCompU{}$, $p = \PosNounCompP{}$, Cohen's $d = \PosNounCompCohenD{}$ raw; $U = \PosNounCorrCompU{}$, $p = \PosNounCorrCompP{}$, $d = \PosNounCorrCompCohenD{}$ after ABTT correction; $d = \StrictNounCorrCohenD{}$ on strictly tangible nouns). (b) Mean cross-lingual orthographic similarity ($1 - \text{NLD}$) confirms that modern international loanwords ($n = \SwadeshCompNumNonSwadesh{}$, e.g., \emph{coffee, radio, internet}) have twice the surface-form overlap ($\OrthoSimLoanword{}$) of both Swadesh core nouns ($\OrthoSimSwadeshNouns{}$) and controlled non-Swadesh nouns ($\OrthoSimControlled{}$).}
  \label{fig:swadesh_comp}
\end{figure*}

Second, we compare Swadesh vocabulary against two 60-item non-Swadesh benchmarks across our \LayerwiseNumLangs{}-language subset (Figure~\ref{fig:swadesh_comp}).
An uncurated non-Swadesh baseline composed of modern international vocabulary (\emph{telephone, radio, coffee, internet, hotel}) exhibits artificially high raw convergence ($\mu = \NonSwadeshCompMean{}$ vs.\ $\mu = \SwadeshCompMean{}$ for all \NumConcepts{} Swadesh items) precisely because recent cultural borrowings share identical or near-identical stems across language families: mean orthographic similarity for the loanword set is $\OrthoSimLoanword{}$, double that of the Swadesh list ($\OrthoSimSwadesh{}$; Figure~\ref{fig:swadesh_comp}b).
Replacing these loanwords with $\ControlledSwadeshCompNumNonSwadesh{}$ non-loanword nouns (\emph{hammer, basket, axe, spider, needle}) lowers non-Swadesh orthographic overlap to $\OrthoSimControlled{}$.
Comparing all $\ControlledSwadeshCompNumSwadesh{}$ Swadesh items against these $\ControlledSwadeshCompNumNonSwadesh{}$ nouns yields a modest difference ($\mu = \ControlledSwadeshCompMean{}$ vs.\ $\ControlledNonSwadeshCompMean{}$; Mann-Whitney $U = \ControlledSwadeshCompU{}$, $p = \ControlledSwadeshCompP{}$, $d = \ControlledSwadeshCompCohenD{}$) because the full Swadesh list mixes concrete nouns with low-convergence grammatical pronouns ($\mu = \CatPronounsMean{}$) and verbs ($\mu = \CatActionsMean{}$).
When we perform an apples-to-apples comparison restricted to \textbf{Swadesh concrete nouns ($n = \PosNounSwadeshCount{}$, orthographic similarity $\OrthoSimSwadeshNouns{}$) versus controlled non-Swadesh nouns ($n = \ControlledSwadeshCompNumNonSwadesh{}$, orthographic similarity $\OrthoSimControlled{}$)} (Figure~\ref{fig:swadesh_comp}a), Swadesh core nouns converge significantly more strongly both in raw embedding space ($\mu = \PosNounSwadeshMean{}$ vs.\ $\PosNounNonSwadeshMean{}$; $U = \PosNounCompU{}$, $p = \PosNounCompP{}$, Cohen's $d = \PosNounCompCohenD{}$) and after ABTT isotropy correction ($\mu = \PosNounCorrSwadeshMean{}$ vs.\ $\PosNounCorrNonSwadeshMean{}$; $U = \PosNounCorrCompU{}$, $p = \PosNounCorrCompP{}$, Cohen's $d = \PosNounCorrCompCohenD{}$).
Furthermore, excluding the 8 semi-abstract or state nouns (\emph{anger, fear, hunger, thirst, dream, harvest, shadow, well}) from the 60 non-Swadesh nouns to form a \textbf{strictly concrete $52$-vs.-$52$ tangible noun comparison} widens the Swadesh advantage even further (raw $\mu = \PosNounSwadeshMean{}$ vs.\ $\StrictNounRawNonSwadeshMean{}$, $U = \StrictNounRawU{}$, $p = \StrictNounRawP{}$, $d = \StrictNounRawCohenD{}$; ABTT $\mu = \PosNounCorrSwadeshMean{}$ vs.\ $\StrictNounCorrNonSwadeshMean{}$, $U = \StrictNounCorrU{}$, $p = \StrictNounCorrP{}$, $d = \StrictNounCorrCohenD{}$).

Third, we probe the \emph{mechanistic source} of this Swadesh noun advantage.
By Zipf's law of abbreviation and cross-linguistic morphological simplexity, culturally universal core nouns are predominantly unanalyzable monomorphemic roots and are significantly shorter across world languages ($\SwadeshNounMeanLen{}$ vs.\ $\NonSwadeshNounMeanLen{}$ characters across our \LayerwiseNumLangs{} languages, $p = \NounLenDiffP{}$), whereas peripheral concrete artifacts (\emph{hammer, basket, spider, needle}) more frequently involve polymorphemic derivation or compounding and are therefore segmented into significantly more SentencePiece subword tokens per word (mean fertility $\SwadeshNounMeanFert{}$ vs.\ $\NonSwadeshNounMeanFert{}$ tokens/word).
In an ANCOVA regressing ABTT convergence across the pooled concrete nouns on the Swadesh indicator, subword fertility, and orthographic overlap, subword fertility strongly predicts cross-lingual convergence ($t = \NounFertAncovaT{}$, $p = \NounFertAncovaP{}$) and fully mediates the direct Swadesh indicator ($p = \NounSwadeshAncovaP{}$).
This reveals a precise mechanistic link between human lexical evolution and neural representation geometry: Zipfian brevity in universal core nouns yields more compact subword representations (closer to single-token atomic lexical entries rather than averaged multi-piece compositions), which NLLB-200 aligns more tightly across languages—whereas \emph{within} the Swadesh core vocabulary, semantic category differences operate independently of subword fertility ($F(7, 91) = \CatPartialFertF{}$, $p = \CatPartialFertP{}$).

\subsection{Genealogical Structure in Language-Pair Distances}
\label{sec:results-phylo}

\begin{figure}[htbp]
  \centering
  \includegraphics[width=\columnwidth]{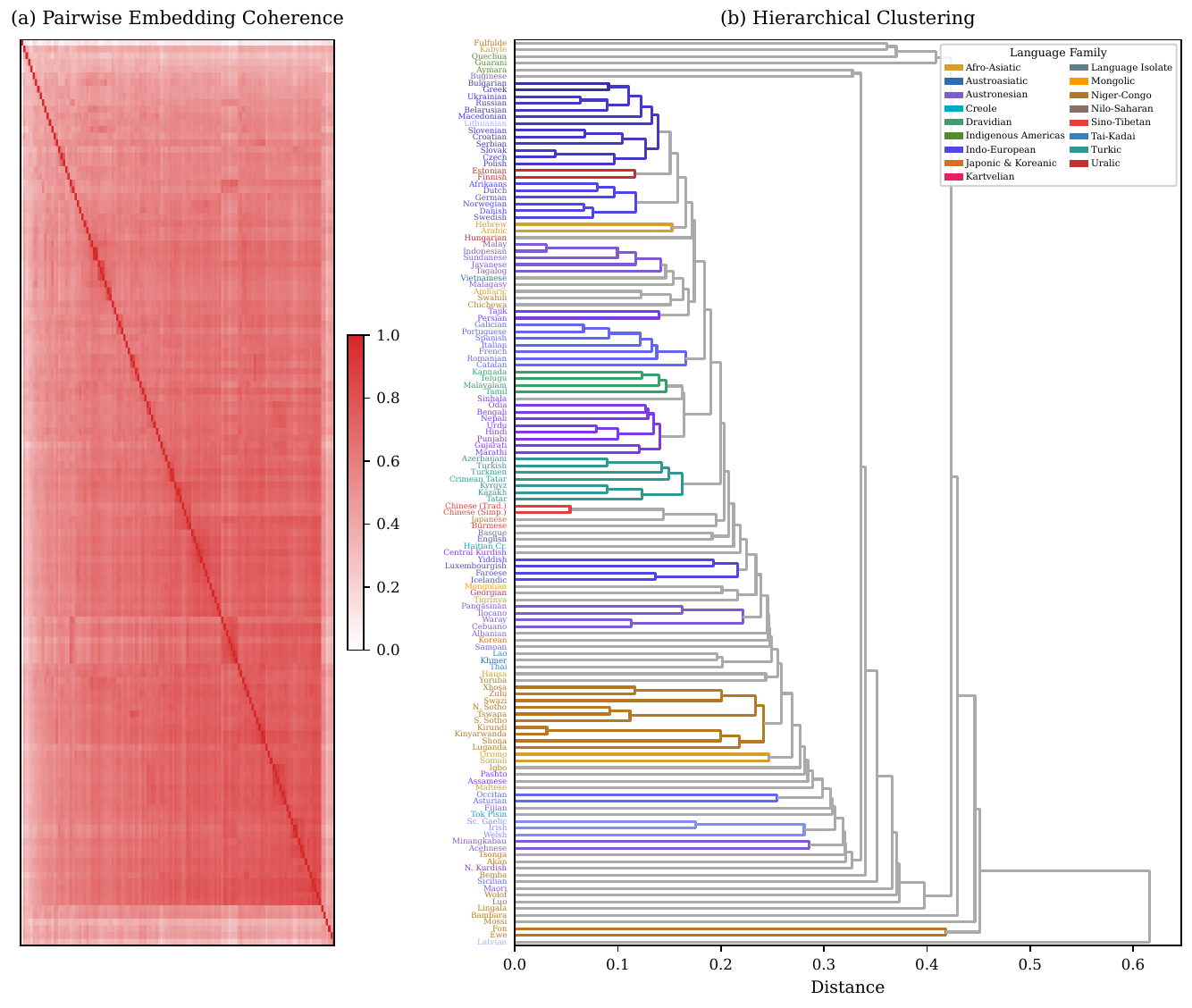}
  \caption{Hierarchical clustering of \PhyloNumLanguages{} languages derived from pairwise NLLB-200 embedding distances averaged across the \NumConcepts{} Swadesh concepts. Major genealogical families and subfamilies (e.g., Romance, Germanic, Slavic, Bantu, Austronesian, Turkic) form contiguous clusters.}
  \label{fig:phylo}
\end{figure}

Having examined concept-level geometry after removing language offsets, we next ask whether the language-pair distances $D_{\text{emb}}(\ell, \ell')$ themselves preserve historical phylogenetic relationships—and whether this signal survives controls for dyadic non-independence and shared writing systems.
Figure~\ref{fig:phylo} displays the pairwise embedding coherence matrix and hierarchical clustering dendrogram across all \PhyloNumLanguages{} languages, revealing cohesive clades for major language families (Indo-European, Austronesian, Niger-Congo, Afro-Asiatic, Turkic, Sino-Tibetan).

\begin{figure}[htbp]
  \centering
  \includegraphics[width=\columnwidth]{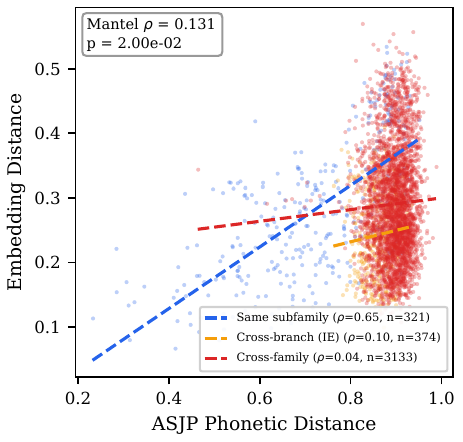}
  \caption{Pairwise NLLB-200 embedding distance vs.\ ASJP lexical phonetic distance across \MantelNumLangs{} languages ($\MantelTotalPairs{}$ pairs), stratified by genealogical relationship tier. Within language subfamilies (blue, $n = \MantelWithinSubfamilyPairs{}$ pairs), embedding distance correlates strongly with phonetic distance ($\rho_s = \MantelWithinSubfamilyRho{}$, block-permutation $p = \MantelWithinSubfamilyBlockP{}$, dyadic $p = \MantelWithinSubfamilyP{}$). Across Indo-European branches (amber, $n = \MantelCrossBranchIEPairs{}$), correlation weakens ($\rho_s = \MantelCrossBranchIERho{}$), and across unrelated language families (red, $n = \MantelBetweenFamilyPairs{}$), ASJP distance saturates near ceiling ($\rho_s = \MantelBetweenFamilyRho{}$). Pooled carrier-frame Mantel $\rho_s = \MantelRho{}$ ($p = \MantelP{}$); decontextualized bare-word Mantel $\rho_s = \MantelBareRho{}$ ($p = \MantelBareP{}$).}
  \label{fig:mantel_scatter}
\end{figure}

To quantify this genealogical correspondence against an external historical benchmark, we compare pairwise embedding distances against ASJP phonetic distances \citep{wichmann2022,jaeger2018} across the \MantelNumLangs{} languages shared by both datasets ($\binom{\MantelNumLangs}{2} = \MantelTotalPairs{}$ pairs; Figure~\ref{fig:mantel_scatter}).

Stratifying language pairs by genealogical depth (Figure~\ref{fig:mantel_scatter}) explains both why NLLB-200 captures local phylogeny and why pooled global Mantel correlations appear modest.
Within language subfamilies ($n = \MantelWithinSubfamilyPairs{}$ pairs across Romance, Germanic, Slavic, Indo-Iranian, Bantu, and Turkic), embedding distance on borrow-resistant Swadesh vocabulary tracks ASJP phonetic divergence with high fidelity ($\rho_s = \MantelWithinSubfamilyRho{}$ with NLD and $\rho_s = \MantelLDNDSubfamilyRho{}$ with \citeauthor{jaeger2018}'s \citeyear{jaeger2018} off-diagonal normalized LDND in the carrier frame; $\rho_s = \MantelBareWithinSubfamilyRho{}$ on bare words).
Because pairwise distances share nodes, we supplement the standard dyadic test ($p = \MantelWithinSubfamilyP{}$) with a within-subfamily block-permutation Mantel test ($9{,}999$ permutations permuting language rows and columns strictly within each subfamily), confirming that the within-subfamily correlation is highly significant ($p = \MantelWithinSubfamilyBlockP{}$ in the carrier frame; $p = \MantelBareWithinSubfamilyBlockP{}$ on bare words).
To rule out the possibility that within-subfamily alignment merely reflects shared orthography or SentencePiece subword overlap, we further isolate the $\MantelCrossScriptSubfamilyPairs{}$ within-subfamily language pairs written in completely different scripts (such as Cyrillic versus Latin Slavic in Russian--Polish and Serbian--Croatian, Devanagari versus Arabic/Persian Indo-Iranian in Hindi--Urdu, and Cyrillic versus Latin Turkic in Kazakh--Turkish).
Across these strictly cross-script subfamily pairs, embedding distance continues to track ASJP phonetic distance strongly both in the carrier frame ($\rho_s = \MantelCrossScriptSubfamilyRho{}$, $p = \MantelCrossScriptSubfamilyP{}$) and on decontextualized bare words ($\rho_s = \MantelBareCrossScriptSubfamilyRho{}$, $p = \MantelBareCrossScriptSubfamilyP{}$), and a partial Spearman Mantel test controlling for shared script across all $\MantelWithinSubfamilyPairs{}$ within-subfamily pairs yields $\rho_{\text{partial}} = \MantelWithinSubfamilyPartialRho{}$.
At deeper Indo-European time depths ($n = \MantelCrossBranchIEPairs{}$ cross-branch pairs, $\sim 4{,}000$--$6{,}000$ years), phonetic erosion reduces the correlation to $\rho_s = \MantelCrossBranchIERho{}$, while across unrelated language families ($n = \MantelBetweenFamilyPairs{}$ pairs, representing $\MantelBetweenFamilyPct{}\%$ of all pairs), ASJP normalized edit distance saturates in a narrow band near $0.90$--$1.00$ ($\rho_s = \MantelBetweenFamilyRho{}$).
Because $\MantelBetweenFamilyPct{}\%$ of all pairs lie in this saturated cross-family regime—and because the shared English carrier template homogenizes language centroids—the pooled Mantel correlation in the carrier frame is $\rho_s = \MantelRho{}$ ($p = \MantelP{}$, \MantelPermutations{} permutations).
When we remove the shared English carrier frame and evaluate language-pair distances on decontextualized native bare words, the global Mantel correlation nearly doubles to $\rho_s = \MantelBareRho{}$ ($p = \MantelBareP{}$), confirming that native lexical embeddings retain substantial genealogical signal.
At the same time, examining each language's mean cosine distance to the rest of the language pool reveals a striking computational parallel to the proficiency/exposure asymmetry of the Revised Hierarchical Model \citep{kroll1994,kroll2010}: high-resource national languages ($n = \RHMHighResCount{}$) cluster significantly closer to the shared cross-lingual conceptual store than lower-resource languages ($n = \RHMLowResCount{}$; mean cosine distance $\mu = \RHMHighResDist{}$ vs.\ $\RHMLowResDist{}$, Mann-Whitney $U = \RHMResourceU{}$, $p = \RHMResourceP{}$).

\subsection{Universal Colexification Structure}
\label{sec:results-colex}

\begin{figure*}[htbp]
  \centering
  \includegraphics[width=\textwidth]{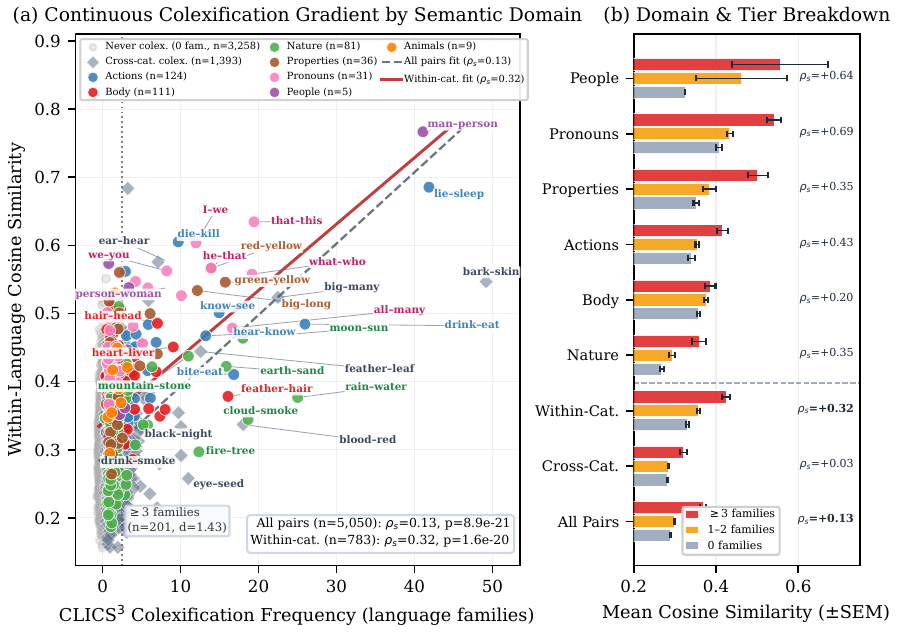}
  \caption{Cross-linguistic colexification structure across all $\binom{\NumConcepts}{2} = \ColexNumPairs{}$ Swadesh concept pairs mapped to CLICS\textsuperscript{3} \citep{list2018,rzymski2020}. (a) Mean within-language concept cosine similarity as a function of colexification frequency (number of distinct language families expressing both concepts with a single word form), stratified by semantic domain: within-category colexified pairs ($n = \ColexWithinCatColexCount{}$) are colored by domain, cross-category colexified pairs ($n = \ColexCrossCatColexCount{}$, e.g., \emph{bark--skin, big--many, ear--hear}) are shown as slate diamonds, and never-colexified pairs ($n = \ColexNonColCount{}$) as grey points. Restricting to within-category pairs ($n = \ColexWithinCatPairs{}$, solid red fit) nearly triples the continuous rank correlation ($\rho_s = \ColexWithinCatRho{}$, $p = \ColexWithinCatRhoP{}$) relative to the unstratified fit ($\rho_s = \ColexSpearmanRho{}$, $p = \ColexSpearmanP{}$, dashed grey fit). (b) Mean cosine similarity ($\pm$SEM) across colexification tiers ($\ge 3$, $1$--$2$, and $0$ families) and semantic domains, showing consistent monotonic ordering within every category.}
  \label{fig:colex}
\end{figure*}

While Section~\ref{sec:results-phylo} tests relationships between languages, our fifth experiment tests whether the fine-grained distances between \emph{concepts} reflect universal human conceptual associations.
We compare pairwise concept cosine similarities against cross-linguistic colexification frequency across all $\binom{\NumConcepts}{2} = \ColexNumPairs{}$ Swadesh concept pairs mapped to CLICS\textsuperscript{3} \citep{list2018,rzymski2020} (Figure~\ref{fig:colex}).

Concept pairs that are frequently colexified across human languages ($\ge 3$ distinct language families, $n = \ColexColCount{}$, such as \emph{bark--skin}, \emph{lie--sleep}, \emph{man--person}, \emph{drink--eat}, \emph{earth--mountain}, \emph{big--many}, and \emph{that--this}) exhibit a mean within-language cosine similarity of $\ColexColMean{}$, compared to $\ColexNonColMean{}$ for concept pairs never colexified in CLICS\textsuperscript{3} ($0$ families, $n = \ColexNonColCount{}$).
A one-sided Mann-Whitney $U$ test confirms a highly significant separation ($U = \ColexU{}$, $p = \ColexP{}$) with a large sample-size-weighted effect size (Cohen's $d = \ColexCohenD{}$), and treating colexification frequency as a continuous variable across all \ColexNumPairs{} pairs yields a positive rank correlation ($\rho_s = \ColexSpearmanRho{}$, $p = \ColexSpearmanP{}$).
Notably, disaggregating across intermediate CLICS\textsuperscript{3} family tiers validates the typological distinction between single-family accidental homophony and recurrent cross-family polysemy \citep{list2018}: pairs attested in only $1$ language family ($n = \ColexOneFamCount{}$, $\mu = \ColexOneFamMean{}$) show virtually no excess similarity over never-colexified pairs ($0$ families, $\mu = \ColexNonColMean{}$), whereas similarity jumps sharply at the CLICS\textsuperscript{3} threshold of $\ge 3$ families ($\mu = \ColexColMean{}$) and climbs monotonically across $\ge 5$ families ($n = \ColexFiveFamCount{}$, $\mu = \ColexFiveFamMean{}$) and $\ge 10$ families ($n = \ColexTenFamCount{}$, $\mu = \ColexTenFamMean{}$).

To ensure that this colexification signal is not a trivial byproduct of identical word strings within our corpus, broad domain or part-of-speech clustering, or the English carrier frame, we subject the comparison to a series of stringent controls.
First, recomputing each concept pair's similarity exclusively across languages where the two concepts are translated by strictly distinct strings ($w_{a,\ell} \neq w_{b,\ell}$) leaves the separation virtually unchanged ($\mu = \ColexStringDistinctColMean{}$ vs.\ $\ColexStringDistinctNonColMean{}$; $U = \ColexStringDistinctU{}$, $p = \ColexStringDistinctP{}$, $d = \ColexStringDistinctCohenD{}$; continuous $\rho_s = \ColexStringDistinctRho{}$, $p = \ColexStringDistinctRhoP{}$), and additionally excluding any language where the two translations are phonetically similar ($\text{NLD}(w_{a,\ell}, w_{b,\ell}) < 0.2$) yields identical results ($U = \ColexCognateExclU{}$, $p = \ColexCognateExclP{}$, $d = \ColexCognateExclCohenD{}$).
Thus, even in languages that lexicalize a concept pair with two completely unrelated words (e.g., English \emph{tree} versus \emph{bark}), NLLB-200 places those concepts significantly closer together if other world languages colexify them.
Second, restricting the string-distinct test strictly to concept pairs within the same semantic category ($n = \ColexWithinCatPairs{}$ pairs) yields an even stronger continuous correlation ($\rho_s = \ColexWithinCatRho{}$, $p = \ColexWithinCatRhoP{}$, nearly tripling the unstratified rank correlation; $\mu = \ColexWithinCatColMean{}$ vs.\ $\ColexWithinCatNonColMean{}$, $U = \ColexWithinCatU{}$, $p = \ColexWithinCatP{}$, $d = \ColexWithinCatCohenD{}$), and restricting to pairs sharing the same part of speech ($n = \ColexSamePOSPairs{}$ pairs) similarly preserves a large effect ($\mu = \ColexSamePOSColMean{}$ vs.\ $\ColexSamePOSNonColMean{}$, $U = \ColexSamePOSU{}$, $p = \ColexSamePOSP{}$, $d = \ColexSamePOSCohenD{}$, $\rho_s = \ColexSamePOSRho{}$).
Third, on language-centered native bare words (excluding identical strings $w_{a,\ell} \neq w_{b,\ell}$), the colexification effect is even stronger than in the carrier frame ($U = \ColexBareStringDistinctU{}$, $p = \ColexBareStringDistinctP{}$, Cohen's $d = \ColexBareStringDistinctCohenD{}$; continuous $\rho_s = \ColexBareStringDistinctRho{}$, $p = \ColexBareStringDistinctRhoP{}$), and evaluating cosine similarity directly between cross-lingual concept centroids $\cos(\hat{\boldsymbol{\mu}}_a, \hat{\boldsymbol{\mu}}_b)$ confirms the same separation in both regimes (carrier $d = \ColexCentroidCohenD{}$, $\rho_s = \ColexCentroidRho{}$, $p = \ColexCentroidRhoP{}$; bare words $d = \ColexBareCentroidCohenD{}$, $\rho_s = \ColexBareCentroidRho{}$, $p = \ColexBareCentroidRhoP{}$).

\subsection{Domain-Level Geometry: Berlin \& Kay Color Hierarchy and Relational Offsets}
\label{sec:results-domains}

In our final set of analyses, we zoom in from the global lexicon to test whether NLLB-200 recovers internal structure within specific semantic domains.

\begin{figure*}[htbp]
  \centering
  \includegraphics[width=\textwidth]{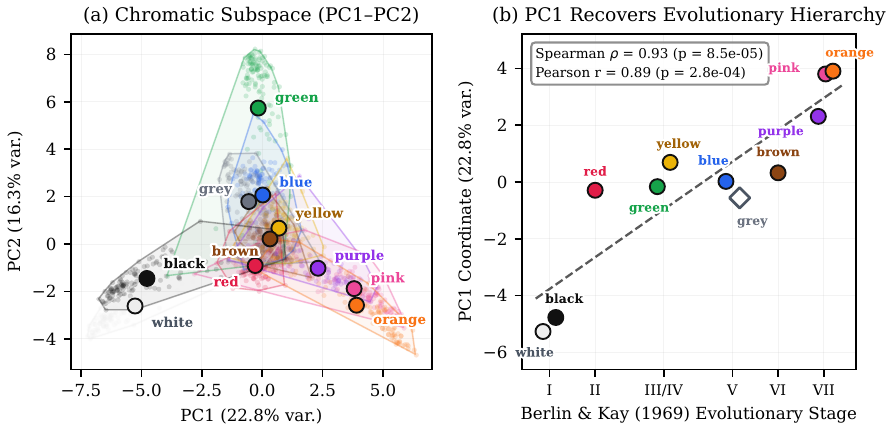}
  \caption{Cross-lingual geometry of the \ColorNumColors{} \citet{berlin1969} basic color terms across \ColorNumLanguages{} languages. (a) Projection onto the PC1--PC2 plane ($\ColorPConeVar{}\% + \ColorPCtwoVar{}\% = \ColorPConeTwoVar{}\%$ of centroid variance): small translucent points denote per-language embeddings, shaded hulls denote cross-lingual dispersion, and large prominent markers denote cross-lingual centroids. PC2 separates warm hues (\emph{red, yellow, orange}) from cool hues (\emph{blue, green}). (b) PC1 coordinate vs.\ \citet{berlin1969} evolutionary stage (Stage~I: \emph{white, black}; Stage~II: \emph{red}; Stages~III--IV: \emph{yellow, green}; Stage~V: \emph{blue}; Stage~VI: \emph{brown}; Stage~VII: \emph{purple, pink, orange}; with \emph{grey} shown as an open diamond at its midpoint stage~$5.0$ representing its wildcard Stage~III--VII status). PC1 directly recovers the typological lexicalization hierarchy ($\rho_s = \ColorBKRho{}$, $p = \ColorBKP{}$; Pearson $r = \ColorBKR{}$, $p = \ColorBKRp{}$).}
  \label{fig:color}
\end{figure*}

\paragraph{Berlin \& Kay Evolutionary Color Hierarchy.}
We extract representations for the \ColorNumColors{} basic color terms of \citet{berlin1969} across \ColorNumLanguages{} languages and apply PCA to their cross-lingual centroids (Figure~\ref{fig:color}).
The top three principal components account for $\ColorTotalVar{}\%$ of the cross-lingual centroid variance ($\text{PC1} = \ColorPConeVar{}\%$, $\text{PC2} = \ColorPCtwoVar{}\%$, $\text{PC3} = \ColorPCthreeVar{}\%$).

Inspection of the principal axes reveals a striking typological regularity.
Rather than encoding a simple physical luminance axis on PC3 (in fact, \emph{black} and \emph{white} share nearly identical coordinates on both PC1 and PC3), \textbf{the primary axis of variance (PC1, $\ColorPConeVar{}\%$) directly orders color terms along the \citet{berlin1969} evolutionary lexicalization hierarchy} (Figure~\ref{fig:color}b): from universally attested Stage~I terms (\emph{white, black}) and Stage~II (\emph{red}), through intermediate Stages~III--VI (\emph{yellow, green, blue, brown}), to late-emerging Stage~VII terms (\emph{purple, pink, orange}).
Across the 10 stage-ordered terms, PC1 correlates with Berlin \& Kay evolutionary stage at Spearman $\rho_s = \ColorBKRho{}$ ($p = \ColorBKP{}$), and across all \ColorNumColors{} terms (assigning \emph{grey} its midpoint stage~$5.0$) at Pearson $r = \ColorBKR{}$ ($p = \ColorBKRp{}$, Spearman $\rho_s = \ColorBKRhoAll{}$, $p = \ColorBKPAll{}$; assigning \emph{grey} Stage~$7.0$ similarly yields $r = \ColorBKRSeven{}$, $p = \ColorBKRSevenP{}$).
Simultaneously, PC2 ($\ColorPCtwoVar{}\%$) separates warm hues (\emph{red, orange, yellow}) from cool hues (\emph{green, blue}) (Figure~\ref{fig:color}a).
Crucially, this evolutionary ordering is not an artifact of the English carrier frame or Zipfian word length: on \textbf{decontextualized native bare-word color embeddings} across all \ColorNumLanguages{} languages, PC1 ($\ColorBarePConeVar{}\%$ of variance) independently recovers the exact same Berlin \& Kay hierarchy ($\rho_s = \ColorBareBKRho{}$, $p = \ColorBareBKP{}$ across the 10 canonical terms; Pearson $r = \ColorBareBKR{}$, $p = \ColorBareBKRp{}$ across all 11 terms), and the partial correlation between PC1 and Berlin \& Kay stage controlling for cross-lingual mean character length remains statistically significant in both the carrier frame ($r_{\text{partial}} = \ColorPartialLengthR{}$, $p = \ColorPartialLengthP{}$) and bare-word regimes ($r_{\text{partial}} = \ColorBarePartialLengthR{}$, $p = \ColorBarePartialLengthP{}$).
Tracing this color manifold across all 12 encoder blocks shows that PC1 stage alignment is already present at Layer~0 ($|\rho_s| = \ColorLayerZeroRho{}$, $p = \ColorLayerZeroP{}$), sharpens to $|\rho_s| = \ColorLayerSevenRho{}$ ($p = \ColorLayerSevenP{}$) by Layer~7, and stabilizes at $\ColorBKRho{}$ at Layer~11, with high-resource languages clustering significantly closer to the cross-lingual color centroids than lower-resource languages ($\mu = \RHMHighResColorDist{}$ vs.\ $\RHMLowResColorDist{}$, $p = \RHMColorP{}$).

\begin{figure*}[htbp]
  \centering
  \includegraphics[width=\textwidth]{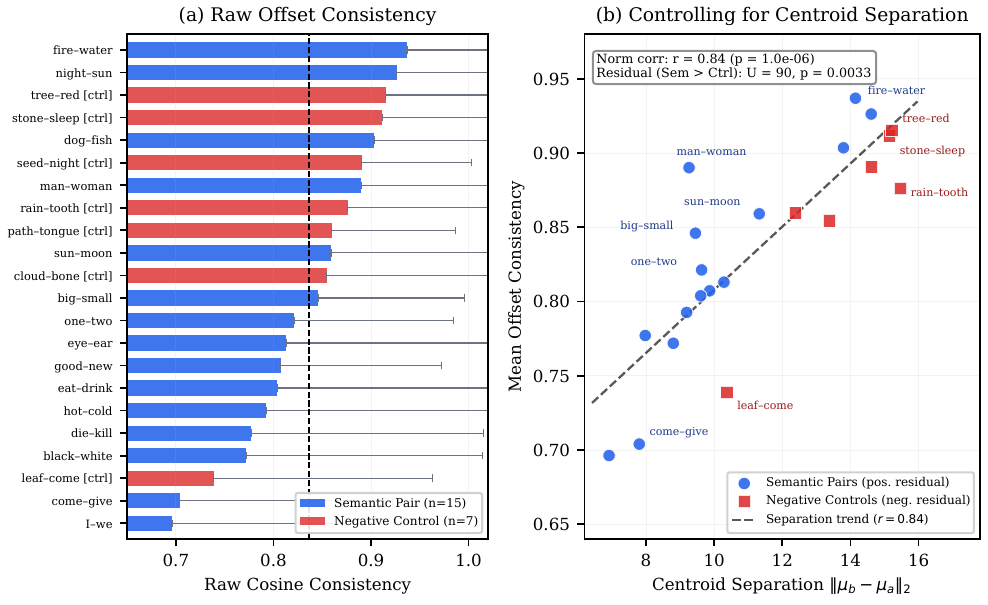}
  \caption{Cross-lingual semantic offset invariance and baseline distance control across \NumLanguages{} languages. (a) Raw offset consistency $S(a,b)$ across \OffsetSemanticCount{} semantic pairs (blue bars, led by \emph{\OffsetBestPair{}} at $\OffsetBestSemanticVal{}$) and \OffsetControlCount{} unrelated cross-domain negative controls (red bars, e.g., \emph{tree--red, stone--sleep, seed--night}). (b) Raw offset consistency correlates strongly with baseline centroid separation $\|\hat{\boldsymbol{\mu}}_b - \hat{\boldsymbol{\mu}}_a\|_2$ ($r = \OffsetNormCorrR{}$, $p = \OffsetNormCorrP{}$) because subtracting two embeddings cancels the shared language vector $\hat{\boldsymbol{\ell}}$. Relative to the OLS regression line, semantic pairs exhibit significantly positive residual consistency ($\mu_{\text{resid}} = +\OffsetResidSemanticMean{}$, led by \emph{\OffsetBestResidPair{}}, \emph{big--small}, and \emph{fire--water}) whereas negative controls fall consistently below the trend ($\mu_{\text{resid}} = \OffsetResidControlMean{}$; Mann-Whitney $U = \OffsetResidU{}$, $p = \OffsetResidP{}$, Cohen's $d = \OffsetResidCohenD{}$).}
  \label{fig:offset}
\end{figure*}

\begin{figure*}[htbp]
  \centering
  \includegraphics[width=\textwidth]{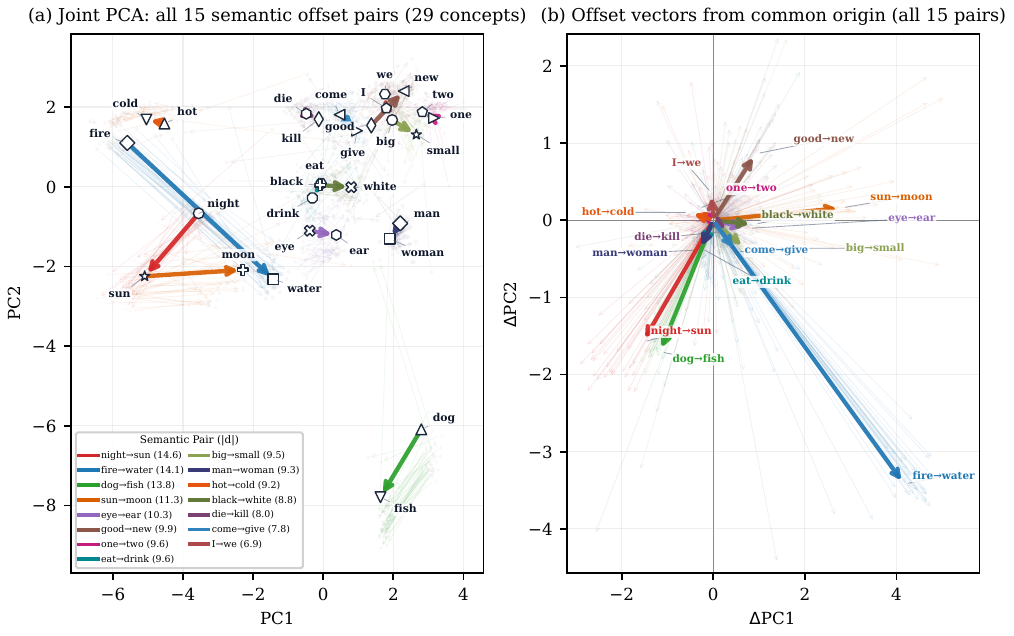}
  \caption{Geometric visualization of cross-lingual offset vectors across all \OffsetSemanticCount{} semantic pairs ($29$ unique concepts) and \NumLanguages{} languages. (a) Joint PCA projection showing per-language embeddings (translucent points), cross-lingual concept centroids (white markers), per-language difference vectors (thin colored arrows), and centroid offset vectors $\hat{\boldsymbol{\mu}}_b - \hat{\boldsymbol{\mu}}_a$ (bold arrows), revealing macro-level clustering into nature, animal, body/person, action, and property/pronoun regions alongside local relational offsets. (b) Per-language and centroid offset vectors translated to a common origin, illustrating directional bundling around each pair's relational axis.}
  \label{fig:offset_vector_demo}
\end{figure*}

\paragraph{Semantic Offset Invariance and Distance-Controlled Residuals.}
Finally, we test whether directional difference vectors $\mathbf{d}_\ell(a,b) = \mathbf{e}_{b,\ell} - \mathbf{e}_{a,\ell}$ remain consistent across \NumLanguages{} languages (Equation~\ref{eq:offset}).
Across all \OffsetNumPairs{} primary pairs (\OffsetSemanticCount{} semantically related antonym/gender/kinship pairs and \OffsetControlCount{} unrelated cross-domain negative controls), raw cross-lingual offset consistency $S(a,b)$ is high ($\mu = \OffsetMeanConsistency{}$, range $\OffsetMinConsistency{}$ to $\OffsetMaxConsistency{}$, led by \emph{\OffsetBestPair{}} at $\OffsetBestSemanticVal{}$; Figure~\ref{fig:offset}a and Figure~\ref{fig:offset_vector_demo}).

However, our additive model (Equation~\ref{eq:intro-decomposition}) exposes a critical geometric baseline that must be controlled in any cross-lingual analogy test.
Because both words in language $\ell$ share the same language shift $\hat{\boldsymbol{\ell}}$, their difference vector $\mathbf{d}_\ell(a,b) = (\hat{\boldsymbol{\mu}}_b - \hat{\boldsymbol{\mu}}_a) + (\hat{\boldsymbol{\epsilon}}_{b,\ell} - \hat{\boldsymbol{\epsilon}}_{a,\ell})$ automatically cancels $\hat{\boldsymbol{\ell}}$.
Whenever the centroid separation $D = \|\hat{\boldsymbol{\mu}}_b - \hat{\boldsymbol{\mu}}_a\|_2$ is large relative to the language-specific perturbations $\hat{\boldsymbol{\epsilon}}$, $\mathbf{d}_\ell(a,b)$ is dominated by the fixed vector $\hat{\boldsymbol{\mu}}_b - \hat{\boldsymbol{\mu}}_a$ even if $\hat{\boldsymbol{\epsilon}}_{a,\ell}$ and $\hat{\boldsymbol{\epsilon}}_{b,\ell}$ are independent.
Indeed, across all \OffsetNumPairs{} pairs, raw consistency $S(a,b)$ correlates strongly with centroid separation $\|\hat{\boldsymbol{\mu}}_b - \hat{\boldsymbol{\mu}}_a\|_2$ ($r = \OffsetNormCorrR{}$, $p = \OffsetNormCorrP{}$; Figure~\ref{fig:offset}b).
Because unrelated cross-domain pairs (\emph{tree--red, stone--sleep, seed--night}) connect distant semantic clusters with large $\|\hat{\boldsymbol{\mu}}_b - \hat{\boldsymbol{\mu}}_a\|_2$, their raw consistency ($\mu = \OffsetControlMean{}$) is mechanically inflated above that of tightly clustered within-domain pairs ($\mu = \OffsetSemanticMean{}$, e.g., \emph{I--we} or \emph{come--give}).

Controlling for centroid separation isolates genuine relational alignment across multiple complementary tests.
Relative to the linear regression trend on $D = \|\hat{\boldsymbol{\mu}}_b - \hat{\boldsymbol{\mu}}_a\|_2$ (Figure~\ref{fig:offset}b), negative control pairs fall systematically below the trend ($\mu_{\text{resid}} = \OffsetResidControlMean{}$), whereas semantically coupled pairs sit above the trend ($\mu_{\text{resid}} = +\OffsetResidSemanticMean{}$, led by \emph{\OffsetBestResidPair{}}, \emph{big--small}, \emph{fire--water}, \emph{sun--moon}, and \emph{one--two}; one-sided Mann-Whitney $U = \OffsetResidU{}$, $p = \OffsetResidP{}$, Cohen's $d = \OffsetResidCohenD{}$).
Using leave-one-language-out centroid scoring ($\cos(\mathbf{d}_\ell, \bar{\mathbf{d}}_{-\ell})$) yields identical separation ($U = \OffsetLOOU{}$, $p = \OffsetLOOP{}$), and fitting the theoretical nonlinear concave baseline $S(a,b) \approx D / \sqrt{D^2 + c}$ likewise confirms significant excess alignment both in the carrier frame ($U = \OffsetConcaveU{}$, $p = \OffsetConcaveP{}$) and on decontextualized bare words ($U = \OffsetBareConcaveU{}$, $p = \OffsetBareConcaveP{}$).
Directly measuring the cosine correlation of the doubly-centered language-specific perturbations $\cos(\hat{\boldsymbol{\epsilon}}_{a,\ell}, \hat{\boldsymbol{\epsilon}}_{b,\ell})$—after subtracting both per-language shifts $\hat{\boldsymbol{\ell}}$ and concept centroids $\hat{\boldsymbol{\mu}}_c$—achieves complete $100\%$ separation between the \OffsetSemanticCount{} semantic pairs and \OffsetControlCount{} negative controls in both the carrier frame ($\mu = +\OffsetCouplingSemanticMean{}$ vs.\ $\OffsetCouplingControlMean{}$, $U = \OffsetCouplingU{}/105$, $p = \OffsetCouplingP{}$, led by \emph{man--woman} $+0.35$, \emph{I--we} $+0.34$, \emph{die--kill} $+0.24$, and \emph{eat--drink} $+0.21$) and the bare-word regime ($\mu = +\OffsetBareCouplingSemMean{}$ vs.\ $\OffsetBareCouplingCtrlMean{}$, $U = \OffsetBareCouplingU{}/105$, $p = \OffsetBareCouplingP{}$): when a language perturbs one member of a semantic pair (e.g., \emph{man} or \emph{hot}), it co-rotates the paired concept (\emph{woman} or \emph{cold}), whereas unrelated concepts vary independently ($\approx 0$).
Finally, comparing the \OffsetSemanticCount{} semantic pairs against \OffsetMatchedCount{} never-colexified negative controls matched on both part of speech and centroid separation ($\|\hat{\boldsymbol{\mu}}_b - \hat{\boldsymbol{\mu}}_a\|_2 = \OffsetMatchedSemNormMean{}$ vs.\ $\OffsetMatchedCtrlNormMean{}$, $p = \OffsetMatchedNormP{}$) shows that semantic pairs achieve significantly higher raw consistency ($\mu = \OffsetMatchedSemMean{}$ vs.\ $\OffsetMatchedCtrlMean{}$, $U = \OffsetMatchedRawU{}$, $p = \OffsetMatchedRawP{}$), residual consistency ($U = \OffsetMatchedResidU{}$, $p = \OffsetMatchedResidP{}$), and perturbation coupling ($U = \OffsetMatchedCouplingU{}/225$, $p = \OffsetMatchedCouplingP{}$), while across languages high-resource languages align more tightly with the cross-lingual centroid offsets than lower-resource languages ($\mu = \RHMHighResOffset{}$ vs.\ $\RHMLowResOffset{}$, $U = \RHMOffsetU{}$, $p = \RHMOffsetP{}$).

\section{Discussion}
\label{sec:discussion}

\subsection{Additive Language--Concept Geometry and Models of the Bilingual Lexicon}

Our central empirical finding is that NLLB-200's shared encoder organizes lexical representations according to an approximately additive decomposition, $\mathbf{e}_{c,\ell} = \bar{\mathbf{e}} + \hat{\boldsymbol{\mu}}_c + \hat{\boldsymbol{\ell}} + \hat{\boldsymbol{\epsilon}}_{c,\ell}$, in which a language-specific translation vector $\hat{\boldsymbol{\ell}}$ is superimposed onto a language-invariant conceptual manifold $\hat{\boldsymbol{\mu}}_c$.
Contrasting our two extraction regimes via two-way sum-of-squares partitioning clarifies why previous evaluations of multilingual encoders have yielded seemingly contradictory estimates of language neutrality \citep{pires2019,conneau2020,chang2022}.
When target words are embedded in a shared English carrier frame, self-attention across the shared context tokens reduces the variance of language shifts $\hat{\boldsymbol{\ell}}$ from $\CSMSSLangBare{}\%$ to $\CSMSSLangCtx{}\%$ before any post-processing, yielding a high raw Conceptual Store Metric ($\text{CSM} = \ConceptualStoreRaw{}$) that improves by $\ConceptualStoreImprovement{}\times$ (to $\ConceptualStoreCentered{}$) after per-language mean-centering.
When native words are embedded in complete isolation without shared context, raw representations are strongly separated by language offset ($\text{CSM} = \ConceptualStoreBareRaw{}$, near the $\CSMPermNull{}$ permutation null), yet subtracting $\hat{\boldsymbol{\ell}}$ produces a $\ConceptualStoreBareImprovement{}\times$ jump in concept separability (to $\ConceptualStoreBareCentered{}$, and $\CSMCrossScriptBareImprovement{}\times$ across strictly disjoint scripts), converging to nearly the same conceptual store ratio as the contextualized regime (Figure~\ref{fig:conceptual_store}).

This additive architecture provides a concrete computational parallel to neurobiological and psycholinguistic models of human bilingualism.
In cognitive neuroscience, \citet{correia2014} used fMRI decoding in bilinguals to localize a language-independent conceptual store in the anterior temporal lobe (ATL), and \citet{deniz2025} recently demonstrated via voxelwise fMRI encoding in Chinese--English bilinguals that cortical semantic representations share a common conceptual geometry modulated both by global language-indicator shifts and by semantic-cluster-specific tuning rotations.
Equation~\ref{eq:intro-decomposition} captures both cortical phenomena simultaneously: our mean-centering operation ($\mathbf{e}_{c,\ell} - \bar{\mathbf{e}}_\ell$) factors out the low-dimensional global language shift $\hat{\boldsymbol{\ell}}$ ($\text{PR} = \PRLangBare{}$) to expose the high-dimensional conceptual manifold $\hat{\boldsymbol{\mu}}_c$ ($\text{PR} = \PRConcept{}$), while the $\sim 30\%$ interaction residual $\hat{\boldsymbol{\epsilon}}_{c,\ell}$ is not unstructured noise but encodes domain-coupled language tuning—as proven by the $100\%$ separation in doubly-centered perturbation co-rotation ($\cos(\hat{\boldsymbol{\epsilon}}_{a,\ell}, \hat{\boldsymbol{\epsilon}}_{b,\ell}) = +\OffsetCouplingSemanticMean{}$ in the carrier frame and $+\OffsetBareCouplingSemMean{}$ on bare words vs.\ $\approx 0$ for controls, $p = \OffsetCouplingP{}$).
In psycholinguistics, the Bilingual Interactive Activation Plus (BIA+) model \citep{dijkstra2002} proposes that visual word input activates representations across all known languages within a single integrated lexicon, while a downstream task-decision system selects the target output language.
NLLB-200 instantiates this division functionally: its shared encoder maps all \NumLanguages{} languages within a single shared parameter space (conditioned only on a prepended source-language tag token rather than separate language-specific encoders), while the decoder's forced target-language BOS token acts as the task-decision gate.
Similarly, our layer-wise trajectory (Figure~\ref{fig:layerwise}) and cross-language dispersion hierarchy mirror the cortical transition from the core fronto-temporal language network \citep{malikmoraleda2022} to downstream conceptual hubs \citep{correia2014} and the Revised Hierarchical Model \citep{kroll1994,kroll2010}: early encoder layers retain lexical and morphosyntactic form ($\mu = \LayerwiseInputConv{}$ at Layer~0), intermediate layers undergo a sharp conceptual phase transition (Layer~\LayerwisePhaseTrans{}), and penultimate layers achieve maximal language-agnostic convergence ($\mu = \LayerwisePeakConv{}$ at Layer~\LayerwisePeakLayer{}) before Layer~11 ($\mu = \LayerwiseFinalConv{}$, centered $\text{CSM} = \LayerwiseFinalCSM{}$) selectively unpacks verb and pronoun morphosyntax for the decoder \citep{voita2019,voita2021}—while across languages, high-resource languages ($n = \RHMHighResCount{}$) exhibit significantly tighter coupling to the shared conceptual store than lower-resource languages ($n = \RHMLowResCount{}$) across Swadesh distances ($p = \RHMResourceP{}$), color centroids ($p = \RHMColorP{}$), and relational offsets ($p = \RHMOffsetP{}$), providing a direct training-exposure analogue to RHM's L1/L2 proficiency asymmetry.

\subsection{Methodological Lessons: Disentangling Universals from Surface and Geometric Confounds}

Beyond its cognitive parallels, our investigation highlights four methodological traps that arise when probing multilingual embedding spaces—and shows how targeted controls resolve them.

\paragraph{Loanword inflation, part-of-speech heterogeneity, and Zipfian subword compactness.}
Comparing core vocabulary against uncurated modern vocabulary (\emph{telephone, radio, coffee, internet}) creates an illusion that non-Swadesh words converge more strongly ($\mu = \NonSwadeshCompMean{}$ vs.\ $\SwadeshCompMean{}$).
Measuring cross-lingual orthographic similarity ($1 - \text{NLD}$) reveals that international loanwords have double the surface-form overlap of core vocabulary ($\OrthoSimLoanword{}$ vs.\ $\OrthoSimSwadesh{}$; Figure~\ref{fig:swadesh_comp}b), allowing shared SentencePiece subwords to bypass semantic abstraction.
Conversely, comparing the entire \NumConcepts{}-item Swadesh list against concrete noun controls conflates lexical stability with part-of-speech heterogeneity because semantic domain explains $\eta^2 = \CatAnovaEtaPct{}\%$ of variance ($F(7,93) = \CatAnovaF{}$, $p = \CatAnovaP{}$; partial $F = \CatPartialFertF{}$, $p = \CatPartialFertP{}$ controlling for subword fertility), with language-specific grammatical deictics and pronouns ranking lowest ($\mu = \CatPronounsMean{}$).
Once matched on part of speech and low orthographic overlap ($\OrthoSimSwadeshNouns{}$ vs.\ $\OrthoSimControlled{}$), Swadesh core concrete nouns converge significantly more strongly than non-Swadesh concrete nouns both before ($p = \PosNounCompP{}$, $d = \PosNounCompCohenD{}$) and after ABTT isotropy correction ($p = \PosNounCorrCompP{}$, $d = \PosNounCorrCompCohenD{}$; $d = \StrictNounCorrCohenD{}$ on strictly tangible nouns; Figure~\ref{fig:swadesh_comp}a)—an advantage mechanistically mediated by Zipfian word-length and subword compactness ($\SwadeshNounMeanFert{}$ vs.\ $\NonSwadeshNounMeanFert{}$ tokens/word; $t = \NounFertAncovaT{}$, $p = \NounFertAncovaP{}$).

\paragraph{Asymptotic saturation of lexical distance and cross-script controls in phylogenetic tests.}
A naive global Mantel test across $\MantelNumLangs{}$ languages yields a modest correlation ($\rho_s = \MantelRho{}$ in the carrier frame; $\rho_s = \MantelBareRho{}$ on bare words).
Stratifying by genealogical tier (Figure~\ref{fig:mantel_scatter}) reveals that this is not a failure of the encoder to represent language family structure, but a known property of ASJP lexical edit distance \citep{jaeger2018,wichmann2022}: phonetic distance tracks historical divergence linearly within language subfamilies ($\rho_s = \MantelWithinSubfamilyRho{}$, block-permutation $p = \MantelWithinSubfamilyBlockP{}$ across $n = \MantelWithinSubfamilyPairs{}$ pairs, and $\rho_s = \MantelCrossScriptSubfamilyRho{}$--$\MantelBareCrossScriptSubfamilyRho{}$ across $n = \MantelCrossScriptSubfamilyPairs{}$ strictly cross-script subfamily pairs), degrades across deep Indo-European branches ($\rho_s = \MantelCrossBranchIERho{}$), and saturates near ceiling across unrelated families ($\rho_s = \MantelBetweenFamilyRho{}$, $\MantelBetweenFamilyPct{}\%$ of all language pairs) where cognate sharing is zero.

\paragraph{Additive language-vector cancellation in relational offset metrics.}
Standard cross-lingual analogy metrics evaluate the directional consistency of $\mathbf{d}_\ell(a,b) = \mathbf{e}_{b,\ell} - \mathbf{e}_{a,\ell}$ without accounting for the fact that subtracting two embeddings from the same language automatically cancels the language shift $\hat{\boldsymbol{\ell}}$.
As a result, raw offset consistency is strongly correlated with baseline centroid separation $\|\hat{\boldsymbol{\mu}}_b - \hat{\boldsymbol{\mu}}_a\|_2$ ($r = \OffsetNormCorrR{}$, $p = \OffsetNormCorrP{}$), inflating distant cross-domain negative controls (\emph{tree--red, stone--sleep, seed--night}).
Controlling for centroid separation via linear or concave regression (Figure~\ref{fig:offset}b), matching controls on distance and part of speech ($p = \OffsetMatchedResidP{}$), or measuring doubly-centered perturbation coupling $\cos(\hat{\boldsymbol{\epsilon}}_{a,\ell}, \hat{\boldsymbol{\epsilon}}_{b,\ell})$ ($+\OffsetCouplingSemanticMean{}$ vs.\ $\OffsetCouplingControlMean{}$, $U = \OffsetCouplingU{}/105$, $p = \OffsetCouplingP{}$) cleanly isolates genuine relational pairs (\emph{man--woman, big--small, fire--water}).

\paragraph{Typological vs.\ perceptual axes in domain subspaces.}
Finally, our analysis of the Berlin \& Kay color space (Figure~\ref{fig:color}) demonstrates the value of quantitative hypothesis testing and bare-word controls over visual inspection of PCA plots.
While the PC1--PC2 plane organizes chromatic terms into warm and cool clusters, the dominant principal axis (PC1, $\ColorPConeVar{}\%$ of variance in the carrier frame; $\ColorBarePConeVar{}\%$ on bare words) is not a physical luminance axis; rather, it orders color concepts monotonically along the \citet{berlin1969} evolutionary lexicalization hierarchy from Stage~I (\emph{white, black}) and Stage~II (\emph{red}) to Stage~VII (\emph{purple, pink, orange}; $\rho_s = \ColorBKRho{}$ in the carrier frame, $\rho_s = \ColorBareBKRho{}$ on bare words, and partial $r = \ColorPartialLengthR{}$ controlling for word length).
Together with the $\ColexNumPairs{}$-pair CLICS\textsuperscript{3} colexification gradient—which filters out single-family accidental homophony ($\mu = \ColexOneFamMean{}$ vs.\ $\ColexColMean{}$ at $\ge 3$ families) and holds across string-distinct ($w_{a,\ell} \neq w_{b,\ell}$, $d = \ColexStringDistinctCohenD{}$), within-category ($\rho_s = \ColexWithinCatRho{}$), and bare-word ($d = \ColexBareStringDistinctCohenD{}$) controls—this demonstrates \emph{latent conceptual colexification}: NLLB-200's continuous geometry encodes the cross-linguistic probability that two concepts can be colexified in human language even when evaluating a language that dislexifies them into distinct words.

\subsection{Limitations}
\label{sec:limitations}

Several methodological limitations qualify our conclusions.
First, regarding contextual framing versus bare-word polysemy, our shared carrier regime splices target words into an English SVO frame (\texttt{"I saw a }$\{$\texttt{word}$\}$\texttt{ near the river."}), creating code-mixed inputs for non-English languages.
While our bare-word baseline eliminates English context entirely and confirms both rank-order preservation ($\rho_s = \CarrierBaselineRho{}$) and a $\ConceptualStoreBareImprovement{}\times$ CSM improvement after mean-centering, isolated bare words lack syntactic disambiguation, which depresses convergence for English-polysemous glosses (\emph{bark, lie, fly}); future benchmarks should average embeddings across native-language parallel sentences in FLORES-200 \citep{nllbteam2022}.
Second, NLLB-200's 256,000-token SentencePiece vocabulary segments words unevenly across writing systems and resource tiers, and mean-pooling subword tokens can introduce representation variance for agglutinative or low-resource scripts, which motivated our $3\times\text{IQR}$ outlier screen excluding six languages.
Third, our experiments probe a single distilled model checkpoint (\texttt{facebook/nllb-200-distilled-600M}); comparing this student model against the 1.3B and 3.3B dense checkpoints and the 54.5B Mixture-of-Experts teacher \citep{nllbteam2022} is needed to determine how model scale and capacity affect conceptual separation.
Finally, NLLB-200 learns exclusively from translation co-occurrence statistics in parallel text without direct visual, auditory, or sensorimotor grounding; the geometric alignment we observe with \citet{berlin1969} color stages, the \emph{Multi-SimLex} concreteness gradient \citep{vulic2020}, and CLICS\textsuperscript{3} colexification demonstrates how biological and perceptual constraints sedimented into the lexical boundaries and distributional translations of the world's languages can be recovered by a purely text-trained encoder \citep{thierry2007,berlin1969,list2018}.

\section{Conclusion}
\label{sec:conclusion}

By probing the shared encoder of NLLB-200 across \NumLanguages{} languages and \NumConcepts{} core concepts under rigorous geometric and typological controls, this study demonstrates that neural machine translation reconciles cross-linguistic diversity with semantic invariance through an approximately additive architecture.
Across the 12-layer encoder stack, representations transition from surface-sensitive early encoder representations at Layer~0 to a language-general conceptual manifold that peaks in cross-lingual alignment at Layer~\LayerwisePeakLayer{} before expanding between-concept separation and unpacking verb/pronoun morphosyntax for decoding at Layer~11.
Within this manifold, subtracting each language's mean vector $\bar{\mathbf{e}}_\ell$ reveals a language-independent conceptual store across both contextualized and bare-word regimes (including across strictly disjoint scripts); semantic category explains over twenty times the variance of surface form while Zipfian subword compactness mediates the convergence advantage of core concrete nouns over peripheral nouns; within-subfamily language distances track historical phonetic divergence across both same-script and cross-script pairs; inter-concept distances mirror universal colexification networks across \ColexNumPairs{} pairs even when restricted to string-distinct and within-category pairs; the primary axis of the color subspace recovers the \citet{berlin1969} evolutionary hierarchy in both contextualized and bare-word regimes; and semantically coupled pairs exhibit coherent relational offsets and doubly-centered perturbation coupling above and beyond baseline centroid separation.

These findings open several concrete directions for future research at the intersection of mechanistic interpretability, linguistic typology, and cognitive neuroscience.
Because our additive decomposition $\mathbf{e}_{c,\ell} = \bar{\mathbf{e}} + \hat{\boldsymbol{\mu}}_c + \hat{\boldsymbol{\ell}} + \hat{\boldsymbol{\epsilon}}_{c,\ell}$ mirrors both the global and cluster-specific language-tuning architecture discovered in bilingual cortex by \citet{deniz2025} and the distinction between the core fronto-temporal language network \citep{malikmoraleda2022} and the anterior temporal lobe conceptual hub \citep{correia2014}, NLLB-200 representations can be used directly as feature spaces in voxelwise fMRI encoding models to test whether early-to-intermediate encoder layers better predict fronto-temporal lexical-syntactic activity while language-centered intermediate-to-penultimate layers (Layers~\LayerwisePhaseTrans{}--\LayerwisePeakLayer{}) better predict cross-lingual conceptual encoding in the ATL and associative semantic cortex.
At the circuit level, extending our layer-wise analysis to individual attention heads and MLP neurons \citep{voita2019,clark2019} across the NLLB-200 scale family (600M, 1.3B, and 3.3B) can localize the specific components responsible for writing and subtracting the language-identity vector $\hat{\boldsymbol{\ell}}$ during forward propagation.
Finally, replacing single-word and template probes with token-aligned representations extracted from full multi-parallel corpora such as FLORES-200 \citep{nllbteam2022} will allow researchers to examine how grammatical categories, morphological case, and context-dependent polysemy interact with universal conceptual geometry across more than 200 languages.

To support cumulative progress and transparent verification, we release the complete \textsc{InterpretCognates} software suite, pre-computed empirical datasets, automated statistical macro pipeline, and interactive 3D visual explorers at \url{https://github.com/kylemath/InterpretCognates} and \url{https://kylemath.github.io/InterpretCognates/}.

\section*{Acknowledgments}
The author thanks Claire Scavuzzo and Alona Fyshe for helpful discussions.

\section*{Data, Code, Interactive Companion, and Programmatic Reproducibility}
To ensure deterministic auditability and eliminate manual transcription errors across all six studies, the complete empirical workflow of \textsc{InterpretCognates} is implemented as an automated, end-to-end pipeline available at \url{https://github.com/kylemath/InterpretCognates}, paired with a synchronized interactive 3D browser companion at \url{https://kylemath.github.io/InterpretCognates/} that follows the exact section and figure progression of this manuscript.
All embeddings are extracted deterministically from the publicly available \texttt{facebook/\allowbreak nllb-200-\allowbreak distilled-600M} checkpoint via Hugging Face \texttt{transformers}, cross-referenced against public typological databases (Swadesh core vocabulary, ASJP v20, and CLICS\textsuperscript{3}), and evaluated through version-controlled Python scripts that automatically generate all 16 publication figures and export all 321 statistical values reported in the text and tables directly as \LaTeX{} macros (\texttt{output/stats.tex}) with zero hardcoded numbers.
In accordance with publisher guidelines on transparent research workflows, the author discloses that AI-assisted coding and editorial tools were utilized under direct human specification to engineer the interactive visualization platform, automate the \LaTeX{} statistical macro verification pipeline, and implement the non-parametric permutation and cross-script control suites; all experimental designs, mathematical formulations, statistical verifications, and scientific interpretations are the sole responsibility of the author.

\bibliographystyle{plainnat}
\bibliography{references}
\end{document}